\documentclass[prepriint,3p,authoryear]{elsarticle}

\makeatletter
\def\ps@pprintTitle{%
 \let\@oddhead\@empty
 \let\@evenhead\@empty
 \let\@oddfoot\@empty
 \let\@evenfoot\@empty}
\makeatother

\usepackage{booktabs}
\usepackage[utf8]{inputenc}
\usepackage{amsmath,amsfonts,amssymb, mathtools, blindtext}
\usepackage{url}
\usepackage{bbold}
\usepackage{wrapfig}
\usepackage{graphicx}
\usepackage{caption}
\usepackage{subcaption}
\usepackage{tabularx}
\usepackage{amssymb}
\usepackage{float}
\usepackage{xcolor}
\usepackage{multirow}
\usepackage{geometry}
\usepackage{threeparttable}
\usepackage{enumitem}
\usepackage[T1]{fontenc}
\usepackage{dsfont}
\usepackage[colorlinks]{hyperref}
\usepackage{bm}

\usepackage[ruled,vlined,linesnumbered]{algorithm2e}
\usepackage{amsmath}
\usepackage{amssymb}
\usepackage{graphicx} % Optional, for including figures

\usepackage[ruled,vlined,linesnumbered]{algorithm2e}
\usepackage{amsmath}
\usepackage{amssymb}
\usepackage{graphicx} % Optional, for including figures

\begin{document}

\begin{frontmatter}

\title{PA-CFL: Privacy-Adaptive Clustered Federated Learning for Transformer-Based Sales Forecasting on Heterogeneous Retail Data}

\author[inst1]{Yunbo Long\corref{cor1}}
% \ead{yl892@cam.ac.uk}
\author[inst1]{Liming Xu}
% \ead{lx249@cam.ac.uk}
\author[inst1]{Ge Zheng}
% \ead{gz305@cam.ac.uk}
\author[inst1,inst2]{Alexandra Brintrup}
% \ead{ab702@cam.ac.uk}
\affiliation[inst1]{
organization={
    Department of Engineering, 
    University of Cambridge, 
    Cambridge},
    country={United Kingdom}
}
\affiliation[inst2]{
    organization={The Alan Turing Institute}, 
    city={London},
    country={United Kingdom}
}

\cortext[cor1]{Corresponding author: yl892@cam.ac.uk}

%% Abstract 
\begin{abstract}
Federated learning (FL) enables retailers to share model parameters for demand forecasting while maintaining privacy. 
However, heterogeneous data across diverse regions, driven by factors such as varying consumer behavior, poses challenges to the effectiveness of federated learning.
To tackle this challenge, we propose Privacy-Adaptive Clustered Federated Learning (PA-CFL) tailored for demand forecasting on heterogeneous retail data.
By leveraging differential privacy and feature importance distribution, PA-CFL groups retailers into distinct ``bubbles'', each forming its own federated learning system to effectively isolate data heterogeneity. 
Within each bubble, Transformer models are designed to predict local sales for each client.
Our experiments demonstrate that PA-CFL significantly surpasses FedAvg and outperforms local learning in demand forecasting performance across all participating clients.
Compared to local learning, PA-CFL achieves a 5.4\% improvement in R\textsuperscript{2}, a 69\% reduction in RMSE, and a 45\% decrease in MAE. 
Our approach enables effective FL through adaptive adjustments to diverse noise levels and the range of clients participating in each bubble. 
By grouping participants and proactively filtering out high-risk clients, PA-CFL mitigates potential threats to the FL system.
The findings demonstrate PA-CFL's ability to enhance federated learning in time series prediction tasks with heterogeneous data, achieving a balance between forecasting accuracy and privacy preservation in retail applications. Additionally, PA-CFL's capability to detect and neutralize poisoned data from clients enhances the system's robustness and reliability.

\end{abstract}

%% keywords

\end{frontmatter}

%% Section: Introduction
\section{Introduction}
The rapid growth of cross-border supply chains and online retail has generated vast amounts of data, enabling the application of machine learning techniques for large-scale demand forecasting \citep{pelaez2024bike}. 
However, challenges such as regional conflicts, trade wars, and data security regulations have made it increasingly difficult for retailers to share privacy-sensitive data across different regions \citep{huang2018retailer,camur2024enhancing}.
Demand data is often decentralized across various stores, regions, or suppliers, and it usually contains sensitive customer information, raising privacy concerns and national security issues \citep{shrestha2020customer}. Furthermore, the volatility of consumer demand across different periods complicates decision-making \citep{bousqaoui2021comparative}, making it difficult for retailers to accurately and efficiently predict demand fluctuations across diverse markets.
Federated learning has emerged as a promising solution to these challenges, as highlighted by \citep{zhong2016big}. As a privacy-preserving approach, FL enables retailers to share model parameters rather than raw data, facilitating collaborative model training while maintaining data security. In the context of demand forecasting, FL provides several advantages. It enhances prediction accuracy for individual retailers, as noted by \citep{li2021demand}, reduces the costs related to global data transfer and storage, and supports real-time model updates through decentralized data. By removing the need to consolidate large datasets, FL improves responsiveness to market fluctuations and boosts operational efficiency.

\begin{figure}[t]
    \centering
\includegraphics[width=\columnwidth]{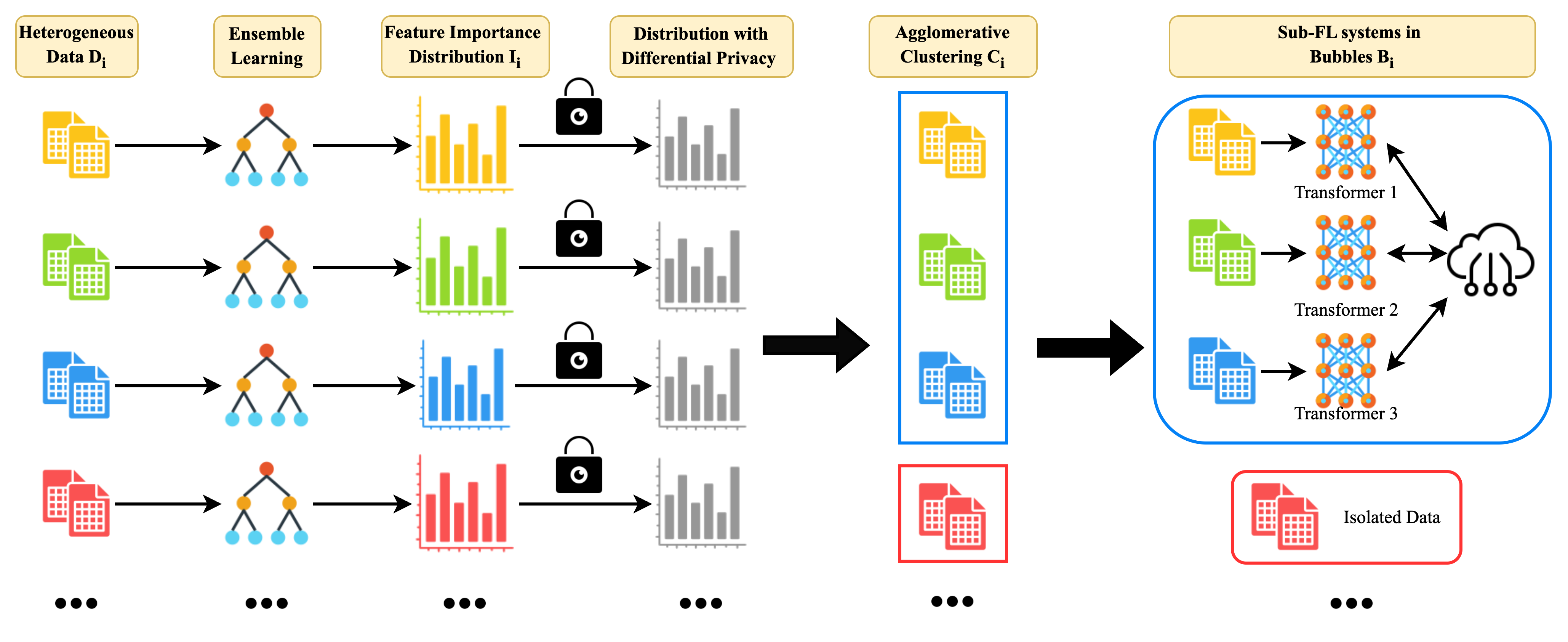}%\rule{6.4cm}{3.6cm}
    \caption{Privacy-Adaptive Clustered Federated Learning framework for heterogeneous data.}
    \label{fig: PFL framework}
\end{figure}

Despite its advantages, FL faces significant challenges in global supply chain applications. Factors such as geographic location, product types, sales seasons, and time-series data scarcity lead to heterogeneous data distributions across regions. As illustrated in \autoref{fig:heterogeneous data}, t-SNE (t-distributed Stochastic Neighbor Embedding) visualizations \citep{van2008visualizing} reveal substantial disparities in demand data across different regions. 
This phenomenon, known as data heterogeneity, poses a major challenge for FL systems in global supply chains. Research has shown that some suppliers fail to benefit from FL, with performance outcomes even worse than those achieved through local learning models \citep{zheng2023federated}. 
These inefficiencies are often caused by the heterogeneous data characteristics of participating suppliers, which can disrupt the global model's performance. 
This highlights a critical limitation of current FL systems as they do not consistently benefit all participants and are vulnerable to the negative impact of heterogeneous data.

To address these challenges, it is essential to develop a more robust FL framework that can intelligently identify suitable retailers for participation while detecting high-risk participants whose data may degrade the model's performance. 
Such a system would facilitate effective collaboration within complex global supply chain, ensuring better outcomes for all stakeholders. 
By mitigating the effects of data heterogeneity and enhancing the resilience of FL systems, this approach can unlock the full potential of federated learning for cross-border retailing demand forecasting.

The main contributions of this paper are summarized as follows:
\begin{itemize}[nosep]
    \item This study proposes a novel clustering-based federated learning framework designed to adjust both the differential privacy noise levels and the number of clusters, allowing flexible management of heterogeneous demand data that could disrupt the federated learning process.
    
    \item Validated on an open-source global retailers dataset, our PA-CFL method outperforms both local learning and the FedAvg (Federated Averaging) method in the demand forecasting, while ensuring that all selected participants can benefit from their respective sub-FL systems.

    \item Through extensive experiments, PA-CFL demonstrates its robustness by flexibly adjusting the differential privacy noise level and the number of participants, guided by the Davies-Bouldin Scores.

\end{itemize}

The rest of this paper is structured as follows:
\autoref{sec:related_work} reviews related work.
\autoref{sec:methodology} introduces the bubble clustering federated learning framework and details the implementation of the PA-CFL algorithm.
\autoref{sec:Experiments_settings} describes the experimental setup and evaluation methods.
\autoref{sec:experiments_results} presents the experimental results and compares the performance of the proposed PA-CFL method with two benchmark approaches through extensive experiments.
\autoref{sec:disscusion} discusses the study's contributions and limitations.
Finally, \autoref{sec:conclusion} provides the conclusion and outlines directions for future research.

%% Section: Related work
\section{Related Work}\label{sec:related_work}
This section reviews the work related to demand forecasting in retail sector, including demand forecasting methods, data sharing, and federated learning approaches.

%% Subsection: Demand Forecasting
\subsection{Demand Forecasting}

Supply chains involve complex data flows, information transfers, and exchanges among various entities, including suppliers, manufacturers, distributors, retailers, and customers. These processes often begin with demand information, which is why much of the research has focused on analyzing downstream retailers' markets to gain comprehensive insights into consumer behavior \citep{yang2021optimal}. Demand forecasting, which predicts future customer demand based on historical sales data, plays a critical role in supply chain management. For online retailers, in particular, accurate forecasting is essential, as it directly impacts the effectiveness of supply chain operations and contributes to profit growth \citep{wisesa2020prediction}.

Quantitative forecasting is widely regarded as one of the most effective approaches for predicting demand and sales prices \citep{kumar2021blockchain}. Among these methods, time series analysis stands out as a primary technique for demand forecasting \citep{prediction}. For instance, quantitative demand forecasting has been successfully applied across various industries, including water resource management \citep{oliveira2017parameter}, rice pricing \citep{ohyver2018arima}, electric vehicle charging predictions \citep{amini2016arima}, and children’s clothing sales forecasting \citep{anggraeni2015performance}. These approaches typically rely on statistical models, such as the Autoregressive Integrated Moving Average (ARIMA), to forecast demand based on historical data collected over time \citep{ramos2015performance}.
In time series analysis, where data evolves sequentially, recurrent neural networks (RNNs) have traditionally been dominant \citep{bandara2019sales}. Variants of RNNs, such as Long Short-Term Memory (LSTM) networks and Gated Recurrent Units (GRUs), have proven particularly effective in capturing temporal dependencies in retail data and adapting to changing conditions \citep{seyedan2020predictive, li2024product}.
While machine learning techniques like RNNs have demonstrated significant value, traditional statistical models often offer greater interpretability \citep{wanchoo2019retail}. This interpretability is crucial in supply chain forecasting, where decision-makers need to understand and trust the factors driving predictions—a key strength of traditional statistical methods \citep{jain2020demand}.

However, as supply chains grow in complexity and the number of features involved in demand forecasting increases, traditional methods often struggle to model these intricate systems effectively \citep{kliestik2022data}. In contrast, machine learning techniques are better suited to capturing the complex patterns and nonlinear relationships inherent in dynamic supply chain demand changes \citep{mediavilla2022review}.
Despite their advantages, deep learning models like LSTM face limitations in processing long-term dependencies efficiently due to issues such as vanishing gradients and sequential data processing \citep{huber2020daily}. In recent years, Transformer models have emerged as a more effective solution for handling time series data \citep{ahmed2023transformers}. Unlike LSTMs, which process data sequentially, Transformers utilize a self-attention mechanism that allows them to focus on the most relevant parts of the data. Additionally, Transformers process time series data in parallel, enabling them to capture long-range dependencies more efficiently \citep{wen2022transformers}. Consequently, Transformer models have gained traction as a promising alternative to traditional demand forecasting methods \citep{oliveira2024evaluating}. When applied to retail demand forecasting, they have consistently demonstrated higher accuracy compared to both traditional statistical methods and earlier machine learning models \citep{ecski2024retail}.

Although machine learning techniques are well-suited for capturing complex patterns and nonlinear relationships in dynamic supply chain demand \citep{mediavilla2022review}, deep learning models like Long Short-Term Memory (LSTM) networks struggle with efficiently processing long-term dependencies due to challenges such as vanishing gradients and the constraints of sequential data processing \citep{huber2020daily}.
In recent years, Transformer models have emerged as a more effective solution for handling time series data,offering improved performance in capturing long-range dependencies \citep{ahmed2023transformers}. 
Unlike LSTMs, which process data sequentially, Transformers utilize a self-attention mechanism that allows them to focus on the most relevant parts of the data. Additionally, Transformers process time series data in parallel, enabling them to capture long-range dependencies more efficiently \citep{wen2022transformers}. As a result, Transformer models have gained traction as a promising alternative to traditional demand forecasting methods \citep{oliveira2024evaluating}. When applied to retail demand forecasting, they have consistently demonstrated higher accuracy compared to both traditional statistical methods and earlier machine learning models \citep{ecski2024retail}.

%% Subsection: Data sharing
\subsection{Data Sharing in Retailers}

Supply chain demand forecasting faces significant challenges due to uncertainty, which can be categorized into internal and external factors. Internally, demand fluctuations caused by shifts in customer preferences, promotional activities, or market trends are difficult for retailers to predict accurately \citep{ren2020demand}. 
Additionally, unexpected demand spikes, particularly during promotions or product launches, can disrupt the accuracy of demand forecasts \citep{datta2011information}. Uncertainties in procurement, production, and shipping lead times further complicate demand forecasting \citep{silva2022demand}. 
Factors such as product life cycles and seasonal variations must also be incorporated into forecasting models. Notably, errors in demand forecasting are often amplified, as illustrated by the bullwhip effect, where inaccuracies propagate along the supply chain, adversely affecting inventory management and production planning \citep{feizabadi2022machine}.
Externally, demand forecasting accuracy is influenced by factors such as economic conditions, geopolitical events, weather, natural disasters, and global health crises \citep{odulaja2023resilience}. These dynamic and unpredictable factors make it challenging to forecast future demand with high accuracy.
To address these uncertainties, suppliers and supply chain partners can collaborate by sharing information to improve forecast accuracy and enhance overall supply chain performance \citep{dai2022two}. For instance, sharing accurate and up-to-date order demand data across different products can enrich the time series data, compensating for data scarcity in specific product categories and improving the overall quality of demand forecasting \citep{hanninen2021information}.
Collaboration can also help mitigate demand fluctuations \citep{noh2020gated} and provide a better understanding of seasonal demand patterns. 
Furthermore, such partnerships enable companies to gain insights into demand in new markets they are entering \citep{abbas2021business}. 
A key driver of supplier collaboration is the recognition of data imbalances between businesses, which can be addressed through shared data to refine demand forecasts across products and geographical regions \citep{alnaggar2021optimization}. By collaborating, suppliers can also better manage external risks, such as raw material shortages or geopolitical disruptions, and mitigate the bullwhip effect \citep{de2015mitigation}.

However, supplier collaboration introduces several challenges. 
A primary concern is data privacy \citep{li2020federated}. Data assets are critical intellectual property, and sharing sensitive information with other suppliers raises significant confidentiality issues \citep{li2019federated}. 
In the retail sector, improper data collection and misuse are common, leading to privacy breaches that may violate regulations like GDPR, especially when sharing data with global companies \citep{borsenberger2022data}.
Another challenge is ensuring fairness in data sharing. Small and medium-sized retailers often partner with large platform operators, such as Amazon, to sell their products. However, these platforms gain access to valuable data, which they can use to advance their own retail businesses, creating an uneven playing field and stifling fair competition \citep{fernandez2022privacy, klimek2021data}. Additionally, direct data sharing reduces the competitive advantage derived from information asymmetry and business secrecy, potentially eroding barriers to competition for both manufacturers and retailers \citep{li2021information}.
To address these challenges, it is crucial to develop a reliable and equitable data-sharing framework for retailers \citep{fernandez2022privacy}. 
This framework should ensure a balance between transparency and security, fostering a collaborative ecosystem that enables mutual benefits while safeguarding competitiveness and privacy.

%% Subsection: Federated Learning in Supply Chain
\subsection{Federated Learning in Supply Chain}

Federated learning \citep{jeong2018communication} is a distributed learning technology designed to enable model training across large-scale, decentralized datasets on multiple devices or servers while preserving data privacy. Instead of sharing raw data, federated learning allows local training on each device or server, with model updates aggregated on a central server to improve the global model \citep{yang2018applied}. This approach has been applied to various supply chain risk management tasks. For instance, in natural gas supply management, federated learning has been used for demand forecasting, enabling natural gas companies and policymakers to develop more accurate supply plans compared to local machine learning approaches \citep{qin2023federated}. Similarly, in the e-commerce sector, federated learning enhances demand prediction models, effectively mitigating the bullwhip effect across the supply chain without requiring direct data exchange \citep{li2021demand}.
And Federated learning is particularly advantageous in scenarios with limited data availability, as sharing model information improves predictive accuracy across participants \citep{kulkarni2020survey}. Knowledge sharing through federated learning reduces overfitting, enhances model performance, and often outperforms traditional machine learning methods \citep{pandiyan2023federated}. For example, federated learning systems based on deep learning models like LSTM have been developed for retail sales forecasting in supply chain contexts \citep{wang2022federated}.

However, these studies face several limitations. Most case studies focus on regional supplier collaboration and do not address the challenges posed by highly diverse supply chain data, such as demand data in global markets. Additionally, research indicates that customers with limited private data benefit the most from federated learning models \citep{kong2024federated}, while the advantages for customers with sufficient data to train their own local models remain unclear \citep{wang2022federated}. Some studies even suggest that not all federated learning participants derive significant benefits \citep{yu2020salvaging}. Furthermore, limited research has explored the use of Transformer models in federated learning for collaborative demand forecasting. As highlighted in \cite{zheng2023federated}, it is essential to ensure that every supplier joins a suitable federated learning system where they can all benefit.
In recent years, trustworthy federated learning \citep{liu2022trustworthy} has emerged, emphasizing privacy protection, fairness, and robustness in federated learning systems. Therefore, there is an urgent need to investigate how to filter suppliers and build a trustworthy federated learning system in a privacy-preserving manner to address inefficiencies in real-world supply chain applications.

%% Subsection: Clustering Federated Learning
\subsection{Clustering Federated Learning}

In real-world training data, such as cross-border e-commerce data, factors like geographic location and product sales often result in non-independent and identically distributed (non-IID) data \citep{zhong2016big}. This non-IID nature poses a significant challenge for federated learning \citep{vahidian2023rethinking}, as it can degrade the performance of the global model in such scenarios \citep{briggs2020federated}.
A common solution to this issue is clustered federated learning (CFL) \citep{ye2023heterogeneous}, which calculates customer similarity based on relevant metrics to facilitate clustering and establish customer selection or grouping strategies. Typically, existing approaches for measuring similarity rely on model weights and local empirical losses \citep{pei2024review}. However, these methods incur significant computational costs when dealing with high model complexity and strong randomness, making it challenging to accurately determine customer similarity \citep{ma2022state}.
Despite its effectiveness in domains like smart grid predictions \citep{chen2022residential}, clustered federated learning has seen limited application in retail demand forecasting within supply chains. This gap highlights the need for further exploration and adaptation of CFL techniques tailored to the unique challenges of retail demand forecasting. Additionally, any implementation of CFL in retail must address customer privacy concerns, ensuring data protection while enabling effective model training and collaboration.

In contrast, grouping strategies based on customer performance or model parameters avoid complex computations but require prior simulation of federated learning to differentiate customers \citep{yan2023clustered}. This approach also becomes computationally intensive with a large number of customers. Given that cross-border retail demand data typically involve numerous features, high data volumes, and a significant number of retail outlets, clustering based on demand data characteristics directly is more efficient.
Current clustering methods, however, may inadvertently lead to privacy breaches \citep{luo2024privacy}. Both model weight clustering and performance-based clustering, derived from simulated federated learning, require sharing model weights and performance metrics with a central computing entity \citep{liao2024predicting,cui2023federated}. 
Moreover, existing CFL emthods are not directly applicable to retail demand forecasting, as retail demand data often exhibits long-range dependencies, where past trends and external events influence future sales over extended periods. Capturing these dependencies is challenging due to noisy fluctuations, missing data, and abrupt demand shifts caused by market dynamics, making traditional approaches ineffective in handling the complexity of retail forecasting.

%% Section: BCFL
\section{The Privacy-Adaptive Clustered Federated Learning}
\label{sec:methodology}

\begin{algorithm}[t!]
\SetAlgoVlined
\small
\caption{Privacy-Adaptive Clustered Federated Learning (PA-CFL)}
\label{alg:PA-CFL}
\textbf{Input:} Local datasets \(\mathcal{D}_i\) for each client \(i\), initial number of clusters \(k\), privacy budget \(\epsilon\), dataset sensitivity of client i $\Delta_i$, learning rate \(\eta\), number of rounds \(T\). \\
\For{each client \(i\)}{
    Train a local XGBoost model on \(\mathcal{D}_i\) to compute feature importance scores \(\mathbf{I}_i\). \\
    Add Laplace noise \(N \sim \text{Laplace}(0, \sigma)\) to \(\mathbf{I}_i\) for differential privacy: \\
    \quad \(\tilde{\mathbf{I}}_i = \mathbf{I}_i + N\), where \(\sigma = \frac{\Delta_i}{\epsilon}\)\\
    Send \(\tilde{\mathbf{I}}_i\) to the central server. \\
}
Aggregate noisy feature importance scores into matrix \(\bm{\tilde{I}} = [\tilde{\mathbf{I}}_1, \tilde{\mathbf{I}}_2, \dots, \tilde{\mathbf{I}}_n]^T\). \\
Normalize each \(\tilde{\mathbf{I}}_i\) to a distribution: \(\tilde{\mathbf{I}}_i \leftarrow \tilde{\mathbf{I}}_i / \sum_j (\tilde{\mathbf{I}}_i)_j\). \\
Perform agglomerative clustering on \(\bm{\tilde{I}}\) using Earth Mover's Distance (EMD): \\
\While{number of clusters \(> 1\)}{
    Compute the distance between clusters: \\
    \quad \(d(\bm{C}_i, \bm{C}_j) = \frac{1}{|\bm{C}_i| \cdot |\bm{C}_j|} \sum_{x \in \bm{C}_i} \sum_{y \in \bm{C}_j} \text{EMD}(x, y)\). \\
}
Determine optimal number of clusters \(k^*\) using Davies-Bouldin Index: \\
\quad \(k^* = \arg\min_{k} DBI(k)\). \\
Assign clients from the same cluster to the each bubble: \(\mathbf{B}_1, \mathbf{B}_2, \dots, \mathbf{B}_{k^*}\). \\

\For{$t=0, \ldots , T - 1$}{
\For{each bubble \(\mathbf{B}_i\)}{
    \If{\(|\mathbf{B}_i| > 1\)}{
        Initialize Transformer model weights \(\mathbf{W}_i(0)\). \\
        \For{each round \(t = 1, \dots, T\)}{
            \For{each client \(j \in \mathbf{B}_i\)}{
                Train local Transformer model on \(\mathcal{D}_j\) with learning rate \(\eta\) to update weights \(\mathbf{W}_j(t)\). \\
                Send \(\mathbf{W}_j(t)\) to the central server. \\
            }
            Aggregate weights using FedAvg: \\
            \quad \(\mathbf{W}_i(t+1) = \frac{1}{|\mathbf{B}_i|} \sum_{j \in \mathbf{B}_i} \mathbf{W}_j(t)\). \\
            Distribute \(\mathbf{W}_i(t+1)\) to all clients in \(\mathbf{B}_i\). \\
        }
    }
    \Else{
        Exclude client \(j\) from federated learning temporarily. \\
    }
}
Compute global model weights: \(\mathbf{W} = \frac{1}{\sum_{i: |\mathbf{B}_i| > 1} |\mathbf{B}_i|} \sum_{i: |\mathbf{B}_i| > 1} |\mathbf{B}_i| \mathbf{W}_i(T)\). \\
Send the global model weights to update the local models $\mathbf{W}_j(t)$}
\textbf{Output:} Global model weights \(\mathbf{W}\), local models weights $\mathbf{W}_j(t)$, clusters \(\mathbf{B}_1, \mathbf{B}_2, \dots, \mathbf{B}_{k^*}\). \\
\end{algorithm}

This section introduces Privacy-Adaptive Clustered Federated Learning (PA-CFL), a novel clustering-based federated learning algorithm inspired by isolation measures used in infectious disease research \citep{kearns2021big}. 
PA-CFL efficiently groups participants into distinct `bubbles' before initiating federated learning, ensuring customer privacy through differential privacy encryption during the grouping process.
The PA-CFL pipeline shown in \autoref{fig: PFL framework} begins with local model training using Gradient Boosting for each participant. Feature importance distributions are then calculated, encrypted, and transmitted to the central server. Using agglomerative clustering, clients are grouped into optimal clusters based on the Davies-Bouldin Score. Each bubble functions as an independent federated learning system, employing a Transformer model for demand prediction. Clients that significantly deviate from others are isolated into their own bubbles, excluded from the federated process, and flagged as potential threats to system stability.
The Privacy-Adaptive Clustered Federated Learning algorithm is presented in \autoref{alg:PA-CFL}, with further details provided in the following subsections.

\subsection{Feature Importance Calculation}  

We apply the extreme gradient boosting (XGBoost) algorithm \citep{chen2016xgboost} to model the joint distribution \( P(\mathbf{X}, Y) \), where \( \mathbf{X} \) represents the input feature matrix and \( Y \) denotes the target variable, effectively capturing complex dependencies to enhance predictive accuracy. After training, \texttt{XGBRegressor} provides insights into feature importance \citep{zheng2017short}, which can be computed based on the contribution of each feature to the overall predictions.  
Let \( \mathbf{I}_{ij} \) denote the importance score matrix for feature \( j \) as computed by client \( i \). This score is calculated by aggregating the impact of feature \( j \) across all splits in the decision trees of the XGBoost model.

\subsection{Differential Privacy}  

Differential privacy \citep{dwork2006differential} provides a framework to protect individual privacy while allowing useful aggregate statistics to be computed. In federated learning, we apply differential privacy to the feature importance distribution calculated by each client to protect sensitive information. This is achieved by adding calibrated noise to the feature importance scores.  
For each client, the local sensitivity is calculated based on its own dataset. Specifically, for client \( i \), the local sensitivity \( \Delta_i \) is defined as  $
\Delta_i = \max_{j, x \in \mathcal{D}_i} \left| \bm{I}_{ij}(\mathcal{D}_i) - \bm{I}_{ij}(\mathcal{D}_i \setminus \{x\}) \right|,$ where \( \mathcal{D}_i \) represents the dataset of client \( i \), \( x \) is a single data point in \( \mathcal{D}_i \), and \( \bm{I}_{ij}(\mathcal{D}_i) \) is the feature importance score for feature \( j \) computed using the full dataset \( \mathcal{D}_i \).  
The feature importance score for feature \( j \) computed after removing the data point \( x \) from \( \mathcal{D}_i \) is represented by \( \bm{I}_{ij}(\mathcal{D}_i \setminus \{x\}) \).  
This formulation ensures that \( \Delta_i \) captures the maximum influence of any single data point \( x \) in client \( i \)'s dataset on the feature importance scores.  

By computing sensitivity locally in this manner, each client can calibrate the noise added to its feature importance scores to provide strong privacy guarantees while preserving the utility of the aggregated statistics in federated learning.  
In differential privacy, the noise scale \( \sigma \) is determined by the sensitivity \( \Delta \) and the privacy budget \( \epsilon \) as follows $ \sigma = \frac{\Delta}{\epsilon}.$ 
A higher \( \epsilon \) indicates less noise and lower privacy protection, while greater sensitivity \( \Delta \) requires more noise to maintain privacy.  
To ensure differential privacy, noise \( N \) is added to the feature importance score. The noise follows a Laplace distribution, which is centered at \( 0 \) with a scale parameter \( \sigma \). The probability density function (PDF) of this noise distribution is given by  
\begin{equation}
    f(N \mid 0, \sigma) = \frac{1}{2\sigma} \exp\left(-\frac{|N|}{\sigma}\right).
\end{equation}  
To generate Laplacian noise, we use a uniform random variable \( U \sim \mathcal{U}\left(-\frac{1}{2}, \frac{1}{2}\right) \), and compute the noise as  
\begin{equation}
    N = -\sigma \cdot \text{sign}(U) \cdot \ln(1 - 2|U|),
\end{equation}  
where \( \text{sign}(U) \) ensures that the noise can take both positive and negative values.  
The differentially private feature importance score for client \( i \) is then computed as $ \tilde{\bm{I}}_{ij} = \bm{I}_{ij} + N,$
where \( \tilde{\bm{I}}_{ij} \) represents the noisy feature importance score that preserves the privacy of client \( i \), and \( N \) is the Laplace-distributed noise added to ensure differential privacy.  
These noisy scores \( \tilde{\bm{I}}_{ij} \) are then transmitted to the central server, where they are used for clustering analysis.

\subsection{Clustering Analysis}

In this stage, we perform agglomerative clustering analysis \citep{mullner2011modern} on the noisy feature importance scores \(\tilde{I}_{ij}\) received from all clients. These scores are organized into a single matrix \(\bm{\tilde{I}}\), where each row corresponds to a client and each column corresponds to a feature. The clustering is performed using Earth Mover's Distance (EMD) \citep{rubner2000earth} as the distance metric. The EMD between two feature importance distributions \(\bm{\tilde{I}}_{j}\) and \(\bm{\tilde{I}}_{j'}\) is defined as follows:
\begin{equation}
d(\bm{\tilde{I}}_{j}, \bm{\tilde{I}}_{j'}) = \min_{\bm{\phi}} \left( \sum_{i=1}^{N} \bm{\phi}(i, j) \cdot c(i, j') \right),
\end{equation}
where \(\bm{\phi}(i, j)\) is the flow of "mass" from point \(i\) in distribution \(\bm{\tilde{I}}_j\) to point \(j'\) in distribution \(\bm{\tilde{I}}_{j'}\), and \(c(i, j')\) is the cost of moving one unit of mass from \(i\) to \(j'\).
Alternatively, EMD can also be expressed in terms of cumulative distribution functions (CDFs) \(F_j\) and \(F_{j'}\):
\begin{equation}
d(\bm{\tilde{I}}_{j}, \bm{\tilde{I}}_{j'}) = \int_0^1 |F_j(x) - F_{j'}(x)| \, dx,
\end{equation}
where \(F_j(x)\) and \(F_{j'}(x)\) are the cumulative distribution functions corresponding to the feature importance vectors \(\bm{\tilde{I}}_j\) and \(\bm{\tilde{I}}_{j'}\), respectively. 
The agglomerative clustering process begins by treating each client as an individual cluster. In each iteration, the two closest clusters are identified based on the cosine similarity distance metric and merged. The distance between two clusters \(\bm{C}_i\) and \(\bm{C}_j\) is determined using the following average linkage criteria:
\begin{equation}
d(\bm{C}_i, \bm{C}_j) = \frac{1}{|\bm{C}_i| \cdot |\bm{C}_j|} \sum_{x \in \bm{C}_i} \sum_{y \in \bm{C}_j} d(x, y),
\end{equation}
where \(d(x, y)\) is the distance between clients \(x\) and \(y\) in terms of their feature importance distributions. This process continues until a predetermined number of clusters \(k\) is achieved or until a stopping criterion is satisfied. 

The resulting clusters are represented as $\bm{Clusters} = \{\bm{C}_1, \bm{C}_2, \ldots, \bm{C}_k\},$ where each cluster \(\bm{C}_i\) contains clients with similar feature importance distributions as $\bm{C}_i = \{j \mid \bm{\tilde{I}}_j \text{ is in cluster } i \}.$
Clients that are isolated or do not fit into any cluster are marked as outliers. 
To select the optimal number of clusters, we use the Davies-Bouldin Index (DBI) \citep{petrovic2006comparison}, defined as follows:
\begin{equation}
DBI(k) = \frac{1}{k} \sum_{i=1}^{k} \max_{j \neq i} \left( \frac{S_i + S_j}{d(\bm{C}_i, \bm{C}_j)} \right),
\end{equation}
where \(S_i\) is the average distance between points in cluster \(\bm{C}_i\), and \(d(\bm{C}_i, \bm{C}_j)\) is the distance between clusters \(\bm{C}_i\) and \(\bm{C}_j\).
The average distance \(S_i\) is computed as 
$$
S_i = \frac{1}{|\bm{C}_i|} \sum_{x \in \bm{C}_i} \sum_{y \in \bm{C}_i} d(x, y)
$$
The optimal number of clusters \(k^*\) is determined by minimizing the Davies-Bouldin Index as $k^* = \arg\min_{k} DBI(k).$
To understand the grouping of clients within these clusters, we define the client-to-cluster assignment function as $G(j) = i \quad \text{if } j \in \bm{C}_i. $
This means that client \(j\) belongs to cluster \(\bm{C}_i\). The final clusters provide insights into the similarity of feature importance distributions across clients, enabling targeted analysis and decision-making in federated learning scenarios.

%% Subsection: Transformer
\subsection{Transformers for Demand Prediction}

\begin{figure}[h] % Use 'htbp' for better placement options
    \centering
    \includegraphics[width=\textwidth]{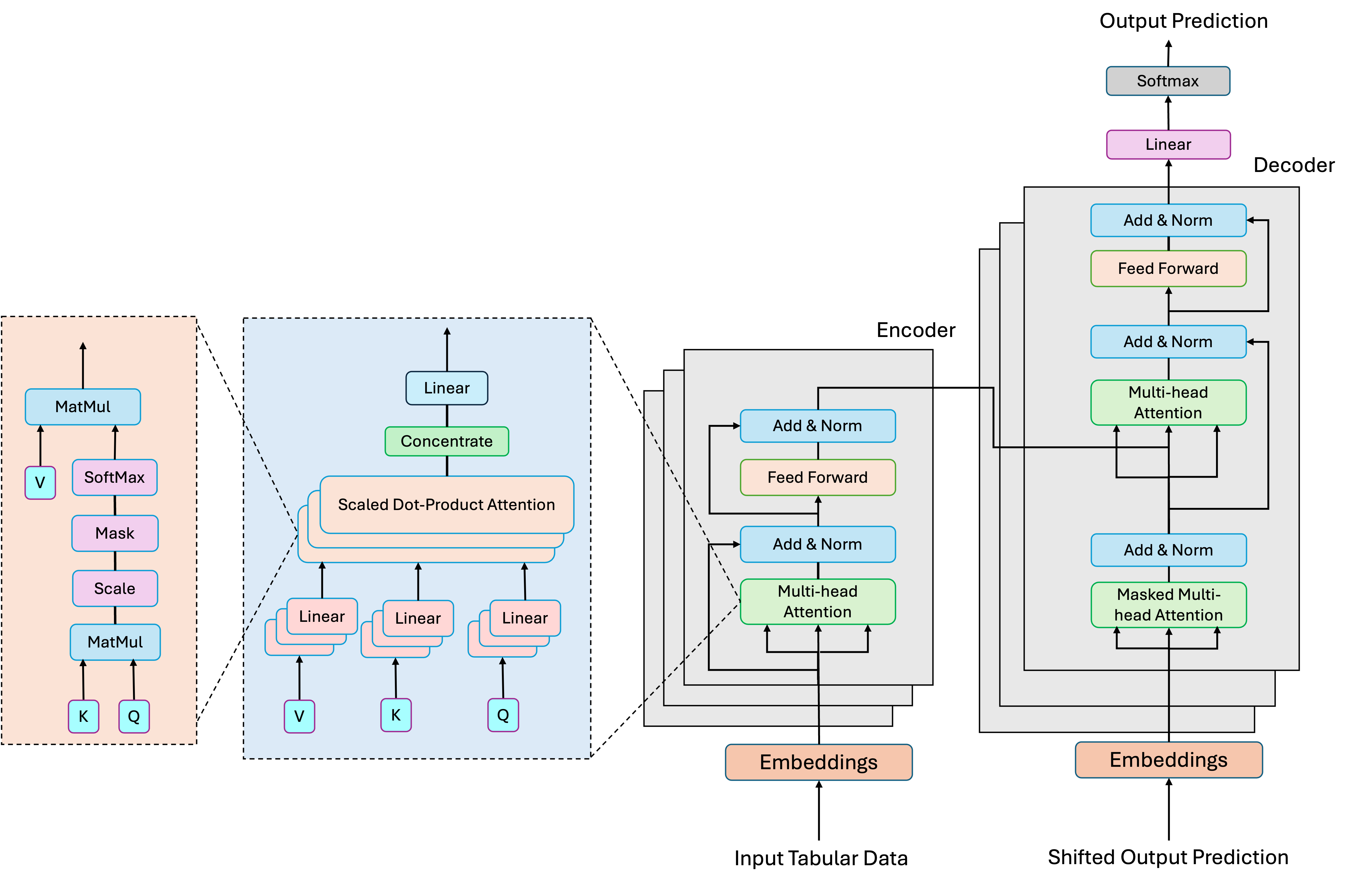}
    \caption{Transformer models for sales prediction.}
    \label{fig:tf}
\end{figure}

Retailers join the federated learning framework to train models based on the Transformer architecture \citep{han2021transformer}, which excels at handling sequential data and time series forecasting, such as demand prediction. We designed a \textit{Sales Transformer Prediction Model}, shown in Figure \ref{fig:tf}, which utilizes a self-attention mechanism to capture dependencies in sequential data. This is particularly useful for demand prediction, where understanding the relationship between past and future demand is critical.
The Transformer architecture consists of an encoder-decoder framework. The encoder transforms input features \(\bm{X} = \{x_1, x_2, \dots, x_T\}\) (e.g., historical sales data) into a continuous representation \(\bm{H} = \{h_1, h_2, \dots, h_T\}\) through embedding layers. This embedding captures the relationships and structures within the data, enabling effective feature representation.
The self-attention mechanism computes attention scores \(\alpha_{ij}\) for each pair of input features \(x_i\) and \(x_j\), allowing the model to dynamically weigh their importance:
\begin{equation}
\alpha_{ij} = \frac{\exp(f(x_i, x_j))}{\sum_{k=1}^{T} \exp(f(x_i, x_k))},
\end{equation}
where \(f(x_i, x_j)\) is computed using the scaled dot-product attention as $f(x_i, x_j) = \frac{\bm{q}_i^\top \bm{k}_j}{\sqrt{d}},$ where \(\bm{q}_i = \bm{W}_Q x_i\) and \(\bm{k}_j = \bm{W}_K x_j\) are the \textit{query} and \textit{key} vectors, respectively, and \(d\) is the dimensionality of these vectors.
These attention scores \(\alpha_{ij}\) are then used to compute weighted sums of the input features as $z_i = \sum_{j=1}^{T} \alpha_{ij} \bm{v}_j,$ where \(\bm{v}_j = \bm{W}_V x_j\) is the \textit{value} vector. This mechanism enhances the model's ability to identify relevant patterns and correlations in demand data.
For demand forecasting, the Transformer model takes historical sales data \(\bm{X}\) as input and predicts future sales \(\bm{Y} = \{y_{T+1}, y_{T+2}, \dots, y_{T+N}\}\). By integrating the Sales Transformer Prediction Model into a federated learning framework, retailers can collaboratively improve demand predictions while preserving data privacy.

\subsection{Federated Learning within Bubbles}

Within each bubble, retailers use their local data to train their models and share their model weights \( \mathbf{W}_k(t) \) with the central server or aggregator at time \( t \). The aggregation process employs the Federated Averaging (FedAvg) method \citep{konevcny2016federated}. Let \( C \) represent the set of all clients, and let \( \mathbf{B}_i \) denote the set of clients in the \( i \)-th bubble. Clients are categorized as follows:

\paragraph{Multi-Client Bubbles} 
If \( |\mathbf{B}_i| > 1 \), where \( |\mathbf{B}_i| \) is the number of clients in bubble \( \mathbf{B}_i \), federated learning is performed among these clients. This allows them to collaboratively improve their models by aggregating their weights. The global model weight update for clients in bubble \( \mathbf{B}_i \) is computed as:
\begin{equation}
\mathbf{W}(t+1)_i = \frac{1}{|\mathbf{B}_i|} \sum_{k \in \mathbf{B}_i} \mathbf{W}_k(t).
\end{equation}

\paragraph{Single-Client Bubbles} 
If \( |\mathbf{B}_j| = 1 \), where \( \mathbf{B}_j \) contains only one client \( j \), this client is flagged as a potential attacker. The rationale is that a single client may lack sufficient data diversity, which could skew the learning process. Such clients are excluded from federated learning. The condition for identifying an attacker is formalized as:
\begin{equation}
\text{Attacker}(j) = \begin{cases} 
1 & \text{if } |\mathbf{B}_j| = 1 \\
0 & \text{if } |\mathbf{B}_j| > 1 
\end{cases}.
\end{equation}

After weight aggregation, the updated global model weights \( \mathbf{W}(t+1) \) are redistributed to each client in the bubbles for the next round of local training. This iterative process continues until the local model training converges. Convergence is assessed using criteria such as the change in the loss function \( \| \mathbf{W}(t+1) - \mathbf{W}(t) \| < k \), where \( k \) is a predefined threshold indicating satisfactory performance.

%% Section: Experimental Settings
\section{Experimental Settings}
\label{sec:Experiments_settings}

\paragraph{Baselines} 
The evaluation of model performance is based on two benchmarks. First, we conduct local learning using sales data from 14 distinct regions. In this setup, each region independently trains its demand forecasting model using only its local data, with no information shared across regions. All regions employ the same Transformer architecture, which is fine-tuned through hyperparameter optimization to maximize prediction accuracy. This benchmark simulates the scenario where retailers in different regions perform Transformer-based demand forecasting independently.
The second benchmark applies the same Transformer architecture for demand forecasting across all regions using the Federated Averaging (FedAvg) algorithm \citep{li2019convergence}, representing a standard FL setup for comparison.
This approach enables the aggregation of performance metrics for each region while facilitating collaborative learning. Both benchmarks are designed to provide a comprehensive comparison of localized and federated learning approaches in the context of demand forecasting.

\paragraph{Experimental Setup} 
To evaluate the effectiveness and robustness of our proposed PA-CFL framework, we conducted a comprehensive set of experiments. The experiments were designed to address three key aspects: comparative performance analysis against baseline methods, sensitivity to privacy parameters, and robustness to clustering configurations.

For the comparative performance analysis, we compared PA-CFL against two baseline approaches: Local Learning and FedAvg. For a fair comparison, we fixed the privacy parameter at \(\epsilon = 10\), representing a medium privacy level, and used the Davies-Bouldin (DB) score to determine the optimal number of clusters. The DB score was chosen due to its ability to balance intra-cluster compactness and inter-cluster separation, ensuring meaningful clustering for federated learning. 

To evaluate the impact of varying privacy levels on PA-CFL performance, we tested three distinct privacy settings: \(\epsilon=0.1\) (high privacy), \(\epsilon=1\) (moderate privacy), and \(\epsilon=10\) (low privacy). These settings were chosen to span a wide range of privacy-preserving scenarios, enabling us to assess the trade-off between privacy and utility. The results were compared against benchmark models to quantify the robustness of PA-CFL under different privacy constraints. 

We further investigated the robustness of PA-CFL under different clustering configurations by varying the number of clusters determined by the DB score. Specifically, we examined whether the system maintains consistent performance across different cluster counts while keeping the privacy level fixed at \(\epsilon = 10\). This analysis ensures that PA-CFL is adaptable to diverse data distributions and clustering outcomes. These experiments collectively demonstrate that PA-CFL provides participants with a flexible and robust framework for balancing data privacy and utility in federated learning systems.

\paragraph{Hyperparameters} 
To ensure optimal performance of the Transformer-based model used in our experiments, we conducted extensive hyperparameter tuning. The model architecture and training parameters were carefully selected to achieve the best results. The Transformer encoder consists of 3 layers, with each layer comprising 8 attention heads in the multi-head attention mechanism. A dropout rate of 0.5 was applied to enhance regularization and prevent overfitting. The model outputs a three-dimensional tensor, with a sequence length of 1 for regression tasks, and a linear layer was used to generate the final regression output. 
For training, we set the learning rate to 0.001, the batch size to 64, and the number of epochs to 50 for local learning and 10 per communication round for federated learning. We employed a grid search strategy to optimize hyperparameters, including learning rate, batch size, and dropout rate. The same initialization parameters, such as weights, biases, and layer configurations, were used for both local and federated learning to ensure consistency. The model was trained iteratively, and prediction accuracy was recorded to select the best-performing hyperparameters. This rigorous tuning process ensures that the model achieves maximum performance while maintaining consistency across different learning scenarios.

\paragraph{Evaluation Metrics} 
To evaluate the performance of the demand regression task, we employed three widely used metrics: R-squared (\(R^2\)), Root Mean Square Error (RMSE), and Mean Absolute Error (MAE). R-squared measures the proportion of variance in the dependent variable that is predictable from the independent variables, providing an indication of the model's goodness of fit. RMSE quantifies the average magnitude of prediction errors and is emphasized in federated learning research due to its sensitivity to large errors. MAE measures the average absolute difference between predicted and actual values and is useful for evaluating the model's robustness to outliers. These metrics collectively provide a comprehensive assessment of the model's predictive accuracy, robustness, and generalization capability.

\paragraph{Hardware and Software} 
All experiments were conducted on a high-performance computing cluster with specific hardware and software configurations. The operating system used was Ubuntu 20.04.6 LTS with a Linux kernel version of 5.15.0-113-generic. The CPU was an Intel(R) Xeon(R) Platinum 8368 processor running at 2.40 GHz, and the GPU was an NVIDIA GeForce RTX 4090 with CUDA support for accelerated deep learning computations. The software stack included Python 3.8, PyTorch 1.12, and TensorFlow 2.10 for model implementation and training. All experiments were repeated 5 times to ensure statistical significance, and the results were averaged to mitigate variability.

%% Section: Experimental evaluation
\section{Experimental Results}\label{sec:experiments_results}

\begin{figure}[t]
    \centering
    \includegraphics[width=0.75\textwidth]{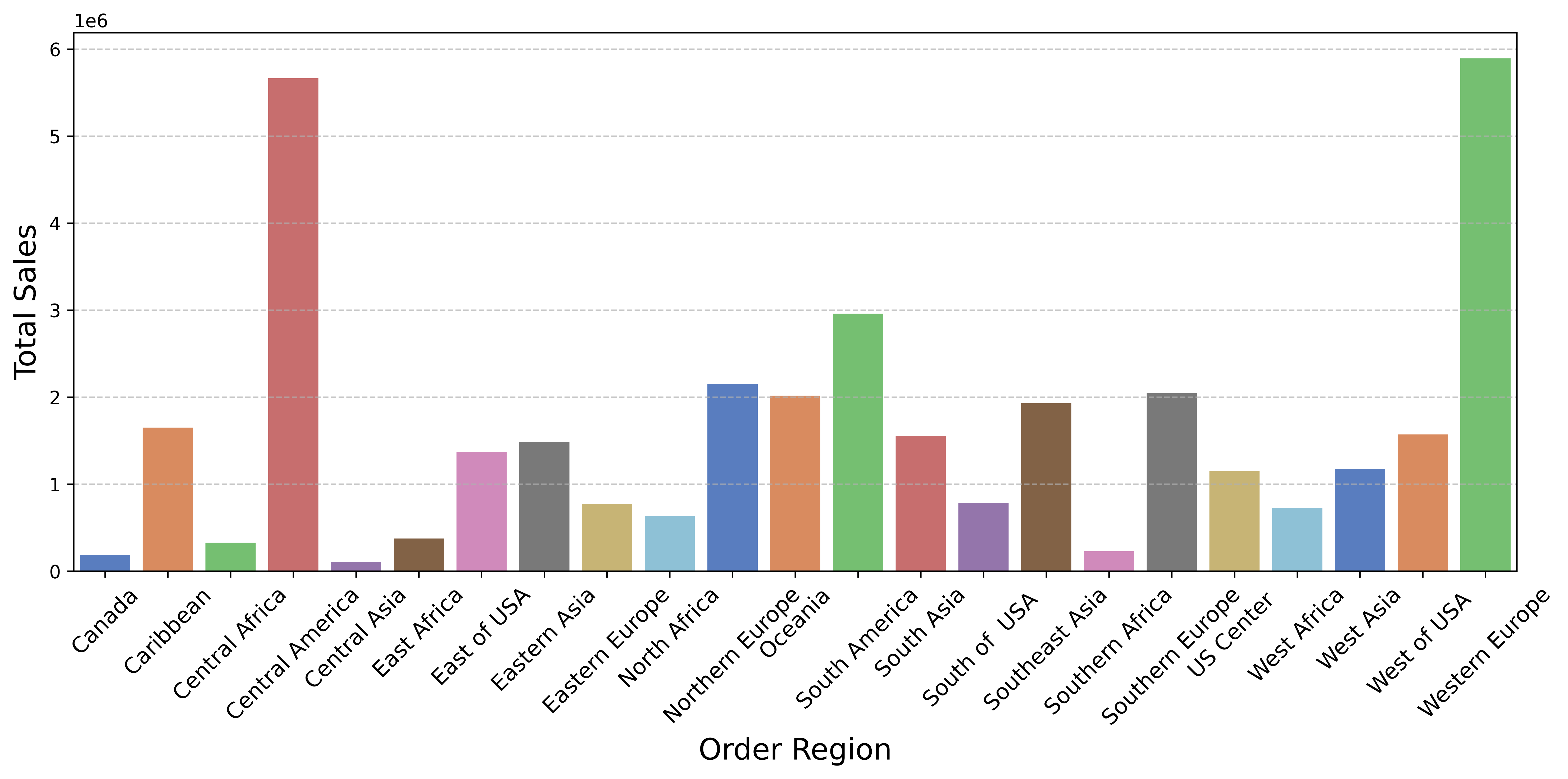}
    \caption{Global sales in different markets.}
    \label{fig: sales_region}
\end{figure}

This section presents the experimental results on demand forecasting using our methods, validated on the DataCo global supply chain dataset \citep{porouhan2021big}. It begins with an exploratory data analysis, followed by a description of data preprocessing and feature engineering for supply chain demand forecasting. Extensive experiments highlight the effectiveness of PA-CFL in handling heterogeneous global retail data.

%% Figure 
\begin{figure}[h]
    \centering
    \includegraphics[width=\textwidth]{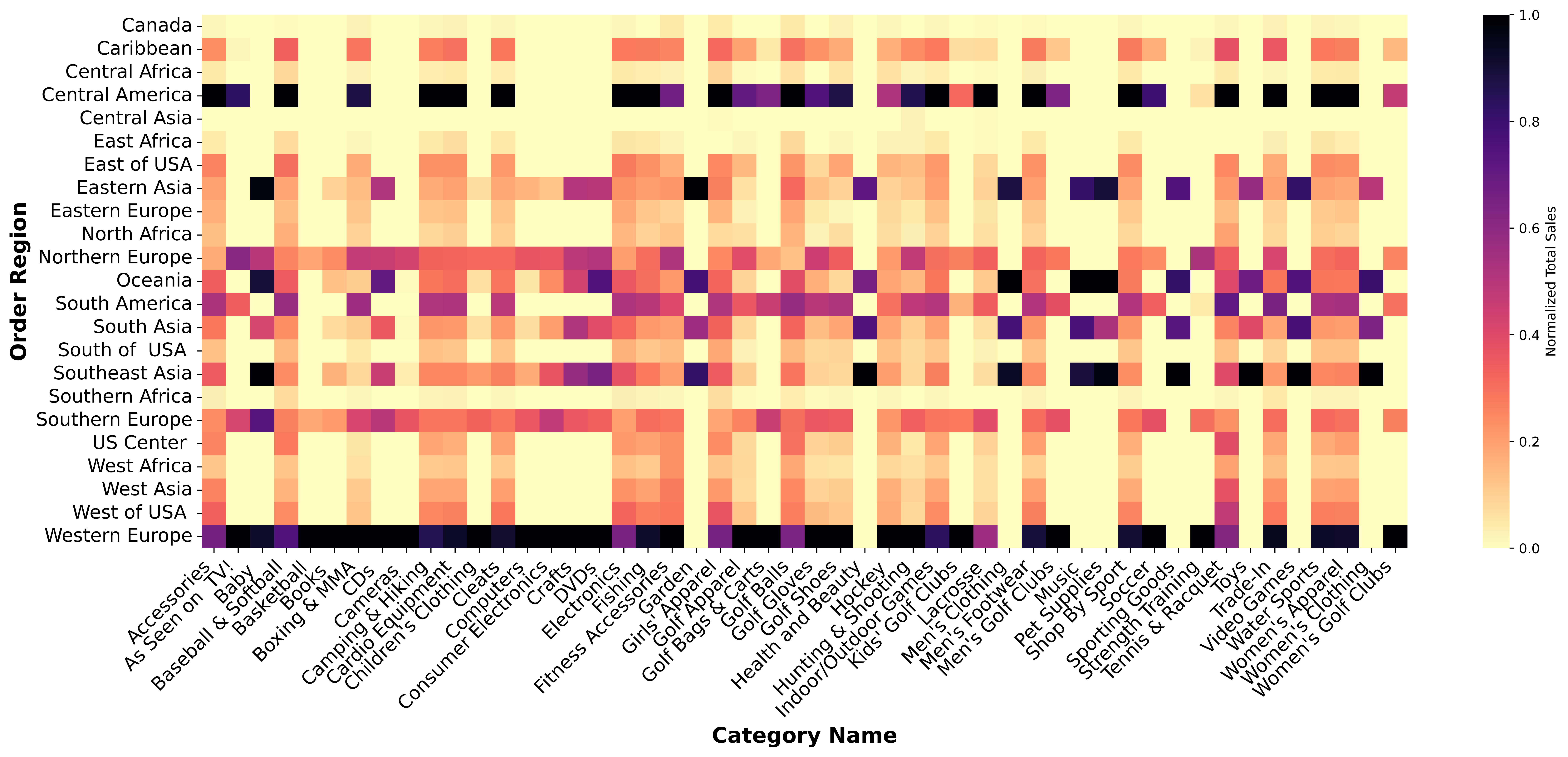}
    \caption{Total sales of different products in diverse order regions}
    \label{fig: sales_product}
\end{figure}

\subsection{Exploratory Analysis}

\begin{figure}[h!]
    \centering
    \includegraphics[width=0.8\textwidth]{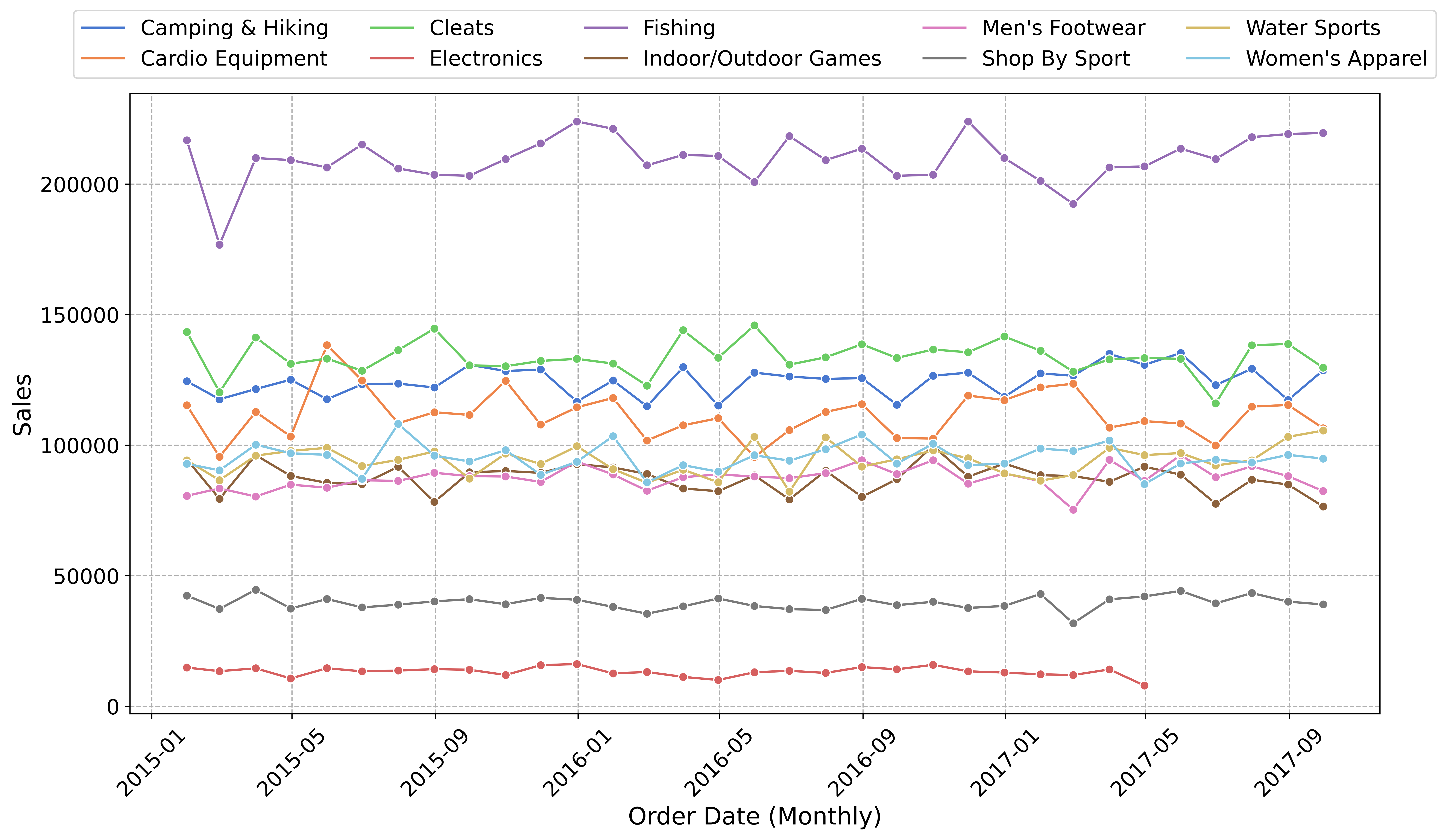}
    \caption{Total sales of different products in diverse order regions}
    \label{fig: sales_TS}
\end{figure}

\begin{figure}[h]
    \centering
    \includegraphics[width=\textwidth]{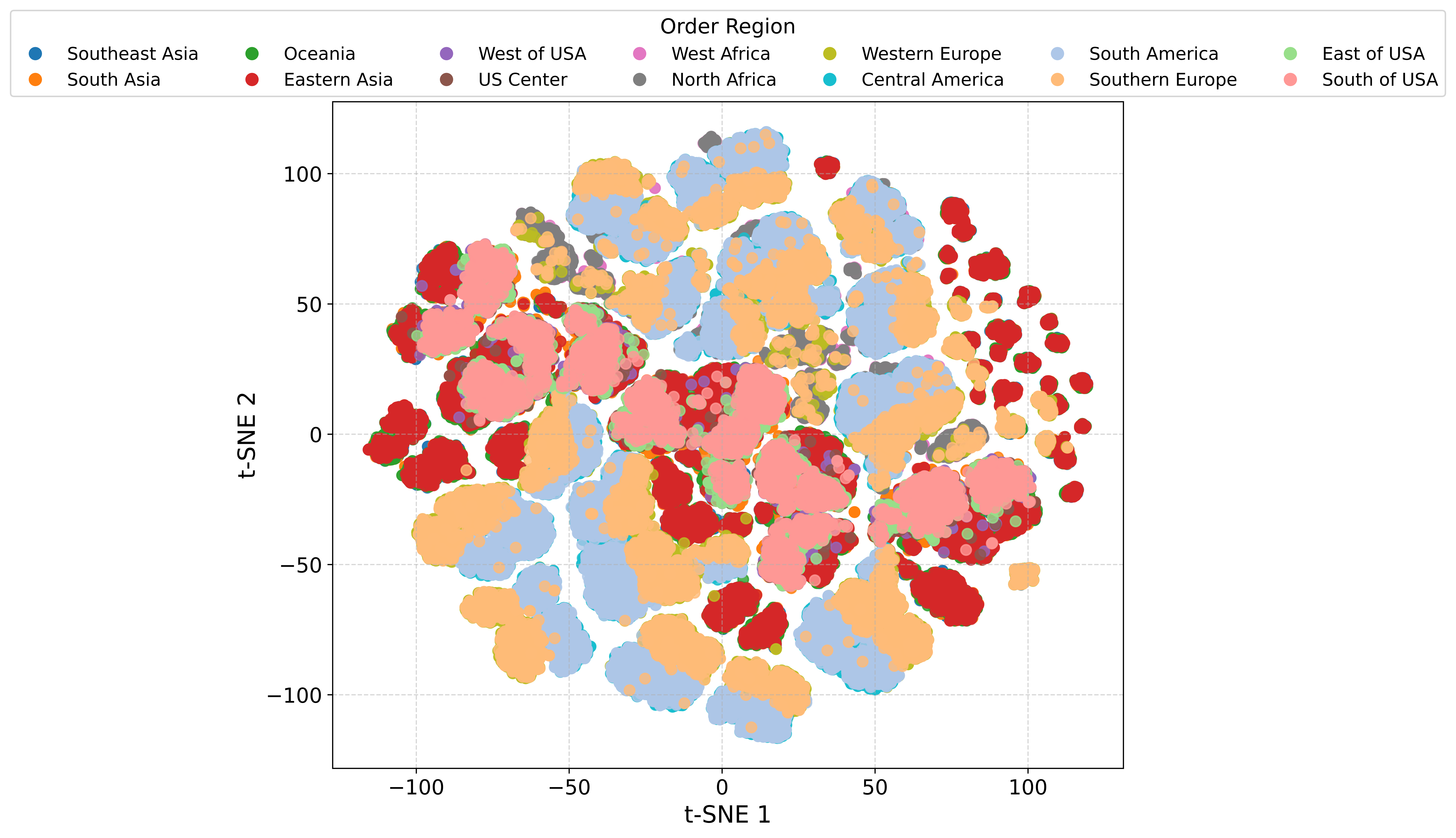} % Adjust width for single-column
    \caption{t-SNE Visualization of Retail Demand Data Heterogeneity}
    \label{fig:heterogeneous data}
\end{figure}

\renewcommand{\arraystretch}{1.2} % Increase \renewcommand{\arraystretch}{1.2} % Increase row spacing
\begin{table}[h!]
    \centering
    \caption{List of districts for participation in federated learning with sensitivity.}
    \footnotesize % Use \footnotesize for smaller font size
    \resizebox{0.85\textwidth}{!}{ % Resize the table to fit the text width
        \begin{tabular}{ccccc} % Added an extra column for Sensitivity
            \toprule
            \textbf{Region} & \textbf{Performance} & \textbf{Quantity of Orders} & \textbf{Continent} & \textbf{Sensitivity} \\ \midrule
            Central America & 0.00032 & 28341 & North America & 0.0185 \\ 
            Western Europe  & 0.00038 & 27109 & Europe & 0.0356 \\ 
            South America    & 0.00071 & 14935 & South America & 0.0227 \\ 
            South Asia       & 0.00088 & 7731 & Asia & 0.0526 \\
            Oceania          & 0.00157 & 10148 & Australia/Oceania & 0.0268 \\ 
            Southeast Asia   & 0.00173 & 9539 & Asia & 0.0651 \\ 
            Eastern Asia     & 0.00182 & 7280 & Asia & 0.0515 \\ 
            West of USA      & 0.00189 & 7993 & North America & 0.0163 \\ 
            Southern Europe  & 0.00257 & 9431 & Europe & 0.0607 \\ 
            East of USA      & 0.00281 & 6915 & North America & 0.0093 \\ 
            South of USA     & 0.00341 & 4045 & North America & 0.0204 \\ 
            US Center        & 0.00440 & 5887 & North America & 0.0195 \\
            West Africa      & 0.00524 & 3696 & Africa & 0.0138 \\ 
            North Africa     & 0.00579 & 3232 & Africa & 0.0184 \\ \bottomrule
        \end{tabular}
    } % End of resizebox
    \label{tab:districts_federated_learning_with_sensitivity}
\end{table}

The dataset includes demand data for various commodities from an e-commerce company across global markets from 2015 to 2018. 
Sales volumes vary significantly by region, as shown in \autoref{fig: sales_region}. 
Western Europe and Central America exhibit the highest sales volumes, exceeding those of most other regions by more than double, whereas Canada and Central Asia report significantly lower sales.
Fishing products dominate global sales, significantly surpassing all other categories (\autoref{fig: sales_product}). Monthly sales data for each product type (\autoref{fig: sales_TS}) reveal diverse sales trends. Compared to sports products, consumer goods for entertainment, such as camping and fishing equipment, exhibit greater volatility.
Additionally, the Retail Demand Data for each region is visualized through a t-SNE projection, as illustrated in \autoref{fig:heterogeneous data}. This visualization highlights significant differences in the distribution of demand data and their associated features, particularly between regions such as South America and Eastern Asia. The distinct clusters observed in the t-SNE projection underscore the heterogeneity present among retailer demand data from various regions, revealing how regional factors contribute to differing demand patterns.

Data filtering focuses on product categories consistently sold from 2015 to 2018, excluding those with insufficient data. Regions are selected based on dataset volume and geographic diversity, with the top 14 regions chosen for federated learning (see ~\autoref{tab:districts_federated_learning_with_sensitivity}). Each region's data is categorized into six features: order, customer, product, supplier, logistics, and payment transaction information. Feature cleaning removes series with excessive missing or incorrect data and eliminates duplicates. Common features across regions are selected for importance ranking and federated training.
The input data comprises 53 feature categories, nearly half non-numeric, converted using one-hot and label encoding. The Pearson correlation coefficient assesses linear relationships between continuous variables, producing a correlation matrix. Time-series characteristics (e.g., order placement, delivery times) are transformed into numeric features (years, months, weeks, days, hours) for feature ranking.
The Pearson Correlation Matrix identifies features strongly correlated with Sales, reducing dimensionality by removing one of two highly correlated non-target features. ANOVA ranks significant features based on F-values and P-values (threshold: 0.06). The top 25 essential features are selected for use by the 14 regional retailers in local, centralized, and federated training.
For the calucaltion of sensitivity, The sensitivity of each region's demand forecasting model is calculated as follows. For each region, an XGBoost regression model is trained on the dataset, with sales figures as the target variable y and other variables as features X. Feature importance is obtained from this model. Each record is then removed iteratively, and the model is retrained to recalculate feature importance. Sensitivity is defined as the maximum change in feature importance due to the removal of any single record. This process is repeated for all records, and the maximum sensitivity for each region is recorded (see \autoref{tab:districts_federated_learning_with_sensitivity}). These sensitivity values are used to apply Laplace noise to each client in the federated learning framework.

\subsection{Demand Forecasting Results}

%% Figure
\begin{figure}[ht]
    \centering
    \begin{subfigure}[b]{0.485\textwidth}
        \includegraphics[width=\textwidth]{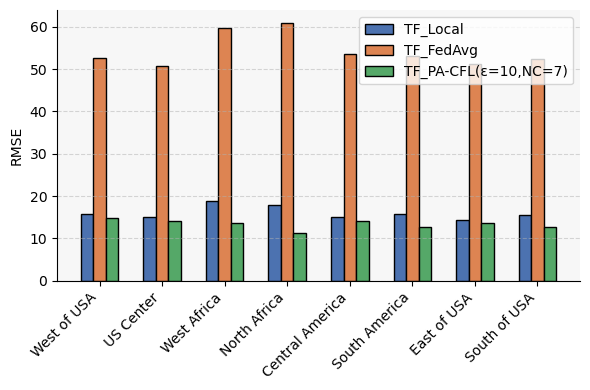}
        \caption{RMSE}
        \label{fig:local1}
    \end{subfigure}
    \begin{subfigure}[b]{0.485\textwidth}
        \includegraphics[width=\textwidth]{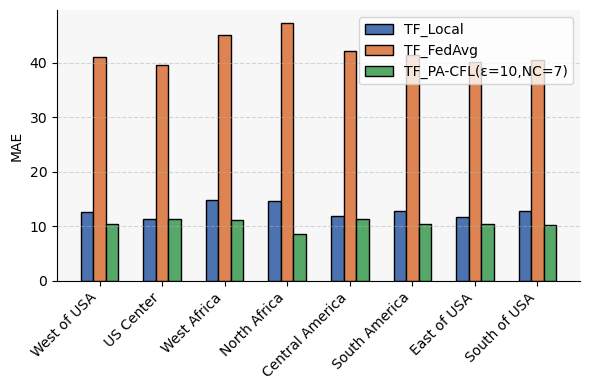}
        \caption{MAE}
        \label{fig:local2}
    \end{subfigure}
    \begin{subfigure}[b]{0.485\textwidth}
        \includegraphics[width=\textwidth]{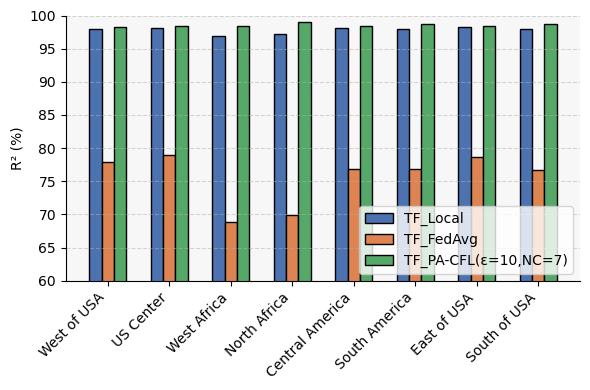}
        \caption{$R^2$}
        \label{fig:local3}
    \end{subfigure}
    \caption{Performance comparison of demand forecasting using Transformers Models (TF) between local learning, FedAvg, and our method, PA-CFL, across three metrics: RMSE, MAE, and $R^2$.}
    \label{fig:local}
\end{figure}

In the first experiment, the performance results of these eight clients under PA-CFL, FedAvg, and local learning are shown in \autoref{fig:local}.
Notably, the outcomes from FedAvg, where all clients participate together, were the least effective. Specifically, the RMSE, MAE, and R² values of the prediction models were significantly worse compared to those achieved through local learning and PA-CFL.
When we compare the performance of PA-CFL with local learning, we observe that PA-CFL consistently yields better results for all participants across various regions, including Africa and America. In particular, the MAE values in PA-CFL are all below 10, which outperforms the values recorded in local learning. This trend is especially evident in North and West Africa, where PA-CFL demonstrates significantly lower RMSE and MAE values compared to local learning, indicating a higher accuracy in testing. Furthermore, PA-CFL achieves \( R^2 \) results that approach 100\%, reflecting a superior model fit. This indicates that PA-CFL not only enhances accuracy but also optimizes the overall performance of the predictive models across diverse geographical regions.

The findings suggest that Privacy-Adaptive Clustered Federated Learning is a more effective approach than FedAvg, particularly in scenarios with dataset heterogeneity \citep{sattler2020clustered}. It also highlights that FedAvg is vulnerable to poisoning by malicious retailers from diverse regions when they input heterogeneous data. Our PA-CFL method addresses this issue by dynamically grouping clients into different clusters based on their input data characteristics at the outset. Additionally, it demonstrates that retailers can benefit from federated learning if appropriate participants are correctly selected to mutually benefit each other.

\subsubsection{Epsilon Values}

% \usepackage{multirow}
% \usepackage{graphicx}
\renewcommand{\arraystretch}{1.5} % Increase row spacing
% \usepackage{multirow}
% \usepackage{graphicx}
% \usepackage{multirow}
% \usepackage{graphicx}
\begin{table}[h]
\centering
\caption{Sales prediction performance comparison among local learning, FedAvg, and Bubble-Clustering Federated Learning (PA-CFL) with varying epsilon values. Lower MAE and RMSE values indicate better performance, while higher \( R^2 \) values reflect improved model fit. The best results are highlighted in {\color{blue}\bf bold}.}
\label{tab:DP}
\resizebox{\textwidth}{!}{%
\begin{tabular}{ccccccccccccccccccc}
\hline
\multirow{3}{*}{\textbf{Region}} &
  \multicolumn{3}{c}{\multirow{2}{*}{\textbf{TF\_Local}}} &
  \multicolumn{3}{c}{\multirow{2}{*}{\textbf{TF\_FedAvg}}} &
  \multicolumn{12}{c}{\textbf{TF\_PA-CFL}} \\ \cline{8-19} 
 &
  \multicolumn{3}{c}{} &
  \multicolumn{3}{c}{} &
  \multicolumn{4}{c}{\textbf{(epsilon=0.1, NC=12)}} &
  \multicolumn{4}{c}{\textbf{(epsilon=1, NC=7)}} &
  \multicolumn{4}{c}{\textbf{(epsilon=10, NC=7)}} \\ \cline{2-19} 
                & RMSE  & MAE   & \(R^2\)      & RMSE   & MAE   & \(R^2\)      & No. & RMSE  & MAE   & \(R^2\)      & No. & RMSE  & MAE   & \(R^2\)      & No. & RMSE  & MAE   & \(R^2\)      \\ \hline
Southeast Asia  & \textcolor{blue}{\bf{31.54}} & \textcolor{blue}{\bf{19.85}} & \textcolor{blue}{\bf{95.05\%}} & 49.29  & 36.38 & 87.93\% & 11  & -     & -     & -       & 1   & -     & -     & -       & 1   & -     & -     & -       \\
South Asia      & \textcolor{blue}{\bf{28.26}} & \textcolor{blue}{\bf{16.64}} & \textcolor{blue}{\bf{95.45\%}} & 44.22  & 34.36 & 88.93\% & 8   & -     & -     & -       & 2   & -     & -     & -       & 2   & -     & -     & -       \\
Oceania         & \textcolor{blue}{\bf{19.12}} & \textcolor{blue}{\bf{13.77}} & \textcolor{blue}{\bf{97.60\% }}& 42.429 & 35.33 & 88.22\% & 1   & -     & -     & -       & 3   & -     & -     & -       & 3   & -     & -     & -      \\
Eastern Asia    & \textcolor{blue}{\bf{39.13}} & \textcolor{blue}{\bf{26.55}} & \textcolor{blue}{\bf{93.53\%}} & 56.04  & 40.20 & 86.77\% & 3   & -     & -     & -       & 4   & -     & -     & -       & 4   & -     & -     & -       \\
West of USA     & 15.74 & 12.56 & 98.01\% & 52.66  & 41.15 & 77.84\% & 4   & \textcolor{blue}{\bf{14.20}} & \textcolor{blue}{\bf{11.59}} & \textcolor{blue}{\bf{98.36\%}} & 5   & 14.90 & 10.52 & 98.22\% & 5   & 14.90 & 10.52 & 98.22\% \\
US Center       & 15.01 & 11.31 & 98.15\% & 50.71  & 39.67 & 78.96\% & 4   & \textcolor{blue}{\bf{9.58}}  & \textcolor{blue}{\bf{7.75}}  & \textcolor{blue}{\bf{ 99.25\%}} & 5   & 14.13 & 11.32 & 98.37\% & 5   & 14.13 & 11.32 & 98.37\% \\
West Africa     & 18.95 & 14.85 & 96.87\% & 59.83  & 45.11 & 68.82\% & 9   & -     & -     & -       & 5   & \textcolor{blue}{\bf{13.63}} & \textcolor{blue}{\bf{11.24}} & \textcolor{blue}{\bf{98.39\%}} & 5   & \textcolor{blue}{\bf{13.63}} & \textcolor{blue}{\bf{11.24}} & \textcolor{blue}{\bf{98.39\%}} \\
North Africa    & 18.01 & 14.72 & 97.28\% & 60.93  & 47.35 & 69.99\% & 5   & -     & -     & -       & 5   & \textcolor{blue}{\bf{11.19 }}& \textcolor{blue}{\bf{8.56}}  & \textcolor{blue}{\bf{98.95\% }}& 5   & \textcolor{blue}{\bf{11.19}} & \textcolor{blue}{\bf{8.56}}  & \textcolor{blue}{\bf{98.95\%}} \\
Western Europe  & \textcolor{blue}{\bf{26.31}} & \textcolor{blue}{\bf{18.54}} & \textcolor{blue}{\bf{97.27\%}} & 74.46  & 48.39 & 78.17\% & 6   & -     & -     & -       & 6   & -     & -     & -       & 6   & -     & -     & -       \\
Central America & 15.19 & 11.88 & 98.14\% & 53.58  & 42.17 & 76.85\% & 10  & -     & -     & -       & 5   & \textcolor{blue}{\bf{14.23}} & \textcolor{blue}{\bf{11.40}} & \textcolor{blue}{\bf{98.37\%}} & 5   & \textcolor{blue}{\bf{14.23}} & \textcolor{blue}{\bf{11.40}} & \textcolor{blue}{\bf{98.37\%}} \\
South America   & 15.88 & 12.76 & 97.93\% & 53.06  & 41.49 & 76.88\% & 7   & 13.52 & 10.82 & 98.50\% & 5   & 12.79 & 10.41 & 98.66\% & 5   & \textcolor{blue}{\bf{12.79}} & \textcolor{blue}{\bf{10.41}} & \textcolor{blue}{\bf{98.66\%}} \\
Southern Europe & \textcolor{blue}{\bf{34.52}} & \textcolor{blue}{\bf{23.63}} & \textcolor{blue}{\bf{95.61\%}} & 77.13  & 49.37 & 78.06\% & 2   & -     & -     & -       & 7   & -     & -     & -       & 7   & -     & -     & -       \\
East of USA     & 14.38 & 11.77 & 98.31\% & 51.26  & 40.17 & 78.71\% & 12  & -     & -     & -       & 5   & \textcolor{blue}{\bf{13.73}} & \textcolor{blue}{\bf{10.44}} & \textcolor{blue}{\bf{98.46\%}} & 5   & \textcolor{blue}{\bf{13.73}} & \textcolor{blue}{\bf{10.44}} & \textcolor{blue}{\bf{98.46\%}} \\
South of USA    & 15.50 & 12.91 & 97.97\% & 52.37  & 40.57 & 76.68\% & 7   & \textcolor{blue}{\bf{10.62}} & \textcolor{blue}{\bf{7.90}}  & \textcolor{blue}{\bf{99.06\%}} & 5   & 12.65 & 10.33 & 98.65\% & 5   & 12.65 & 10.33 & 98.65\% \\ \hline
\end{tabular}%
}
\end{table}

In the PA-CFL framework, the optimal number of clusters for client grouping is determined using the lowest Davies-Bouldin score. Additionally, it is crucial to evaluate how variations in encryption levels, achieved through differential privacy, impact the system's robustness. \autoref{tab:DP} presents the performance of PA-CFL with varying epsilon values (0.1, 1, and 10) and their corresponding number of clusters (NC), representing different levels of privacy preservation.
Besides, it shows the specific group number of the clustering for each region.
The results demonstrate that PA-CFL consistently achieves higher accuracy than both local learning and FedAvg. For instance, in the US Center and South of the USA, the R² values reach 99.25\% and 99.06\%, respectively, significantly outperforming other methods. Furthermore, the MAE values drop to 7.75 and 7.9 when epsilon is set to 0.1, indicating high accuracy even under strong encryption. This highlights the robustness of the PA-CFL model under stringent privacy conditions. Notably, for epsilon values of 1 and 10, performance remains consistent and continues to surpass local learning for all retailers. Increasing epsilon beyond these values does not further improve performance, suggesting that PA-CFL maintains stable robustness across varying encryption levels.
Moreover, as epsilon decreases (implying stronger privacy guarantees), the number of clusters increases, and fewer clients are grouped into a single cluster for federated learning. This is likely because higher privacy preservation makes it more challenging for PA-CFL to accurately group clients with similar data distributions. As a result, PA-CFL selects fewer clients with higher confidence for federated learning, ensuring system reliability despite privacy constraints.
The results demonstrate that, under different encryption levels, our PA-CFL method consistently and effectively groups retailers by determining the optimal number of clusters, as indicated by the Davies-Bouldin Index. A lower Davies-Bouldin score signifies more compact and well-separated clusters, which is essential for accurate client segmentation.
\autoref{fig:score} shows that higher epsilon values consistently yield lower Davies-Bouldin scores, indicating that increased epsilon improves client clustering. This flexibility enables retail participants to encrypt their data at varying privacy levels while maintaining strong performance within the federated learning system. As a result, PA-CFL not only adapts to diverse privacy requirements but also ensures robust learning outcomes across regions and clients.

\subsubsection{Clustering Numbers}

\begin{figure}[t] 
    \centering \includegraphics[width=0.55\textwidth]{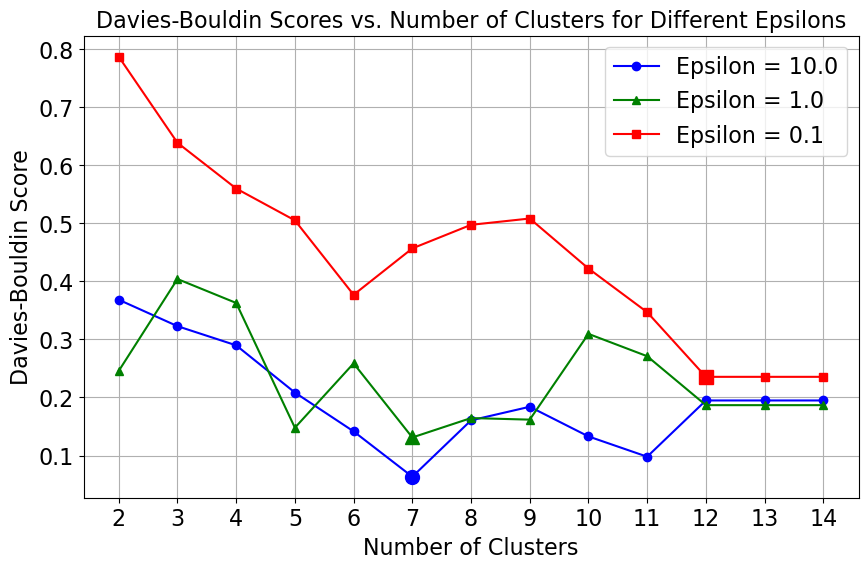} 
    \caption{Davies-Bouldin Index}\label{fig:score}
\end{figure}

\renewcommand{\arraystretch}{1} % Increase row spacing

\begin{table}[ht!]
\centering
\caption{Bubble-Clustering Federated Learning with different number of clustering, ranging from 2 to 7 (Epsilon = 10)}
\label{tab:clustering1}
\resizebox{\textwidth}{!}{%
\begin{tabular}{cccccccccccccccc}
\toprule
\multirow{2}{*}{\textbf{Regions}} &
  \multirow{2}{*}{\textbf{Metrics}} &
  \multirow{2}{*}{\textbf{TF\_Local}} &
  \multirow{2}{*}{\textbf{TF\_FedAvg}} &
  \multicolumn{12}{c}{\textbf{TF\_PA-CFL}} \\ 
\cmidrule(lr){5-16} % Corrected column rule placement
  \multicolumn{2}{c}{} &  % Empty placeholders for alignment
  & & 
  \multicolumn{2}{c}{\textbf{NC =2}} &
  \multicolumn{2}{c}{\textbf{NC =3}} &
  \multicolumn{2}{c}{\textbf{NC =4}} &
  \multicolumn{2}{c}{\textbf{NC =5}} &
  \multicolumn{2}{c}{\textbf{NC =6}} &
  \multicolumn{2}{c}{\textbf{NC =7}} \\ 
\midrule

  % \midrule
\multirow{3}{*}{\textbf{Southeast Asia}} &
  RMSE &
  31.54 &
  49.29 &
  \multirow{3}{*}{1} &
  \textcolor{blue}{\bf{15.65}} &
  \multirow{3}{*}{1} &
  \textcolor{blue}{\bf{15.65}} &
  \multirow{3}{*}{1} &
  \textcolor{blue}{\bf{15.65}} &
  \multirow{3}{*}{1} &
  - &
  \multirow{3}{*}{1} &
  - &
  \multirow{3}{*}{1} &
  - \\
 &
  \(R^2\) &
  95.05\% &
  28.26\% &
   &
  \textcolor{blue}{\bf{98.68\%}} &
   &
  \textcolor{blue}{\bf{98.68\%}} &
   &
  \textcolor{blue}{\bf{98.68\%}} &
   &
  - &
   &
  - &
   &
  - \\
 &
  MAE &
  25.60 &
  36.38 &
   &
  \textcolor{blue}{\bf{12.12}} &
   &
  \textcolor{blue}{\bf{12.12}} &
   &
  \textcolor{blue}{\bf{12.12}} &
   &
  - &
   &
  - &
   &
  - \\ \hline
\multirow{3}{*}{\textbf{South Asia}} &
  RMSE &
  28.26 &
  44.22 &
  \multirow{3}{*}{2} &
  48.42 &
  \multirow{3}{*}{2} &
  \textcolor{blue}{\bf{26.31}} &
  \multirow{3}{*}{2} &
  \textcolor{blue}{\bf{26.31}} &
  \multirow{3}{*}{3} &
  \textcolor{blue}{\bf{26.31}} &
  \multirow{3}{*}{4} &
  - &
  \multirow{3}{*}{5} &
  - \\
 &
  \(R^2\) &
  95.45\% &
  88.93\% &
   &
  85.78\% &
   &
  \textcolor{blue}{\bf{96.12\%}} &
   &
  \textcolor{blue}{\bf{96.12\%}} &
   &
  \textcolor{blue}{\bf{96.12\%}} &
   &
  - &
   &
  - \\
 &
  MAE &
  23.09 &
  34.36 &
   &
  37.67 &
   &
  \textcolor{blue}{\bf{21.01}} &
   &
  \textcolor{blue}{\bf{21.01}} &
   &
  \textcolor{blue}{\bf{21.01}} &
   &
  - &
   &
  - \\ \hline
\multirow{3}{*}{\textbf{Oceania}} &
  RMSE &
  \textcolor{blue}{\bf{19.12}} &
  42.429 &
  \multirow{3}{*}{2} &
  51.24 &
  \multirow{3}{*}{3} &
  67.556 &
  \multirow{3}{*}{4} &
  - &
  \multirow{3}{*}{5} &
  - &
  \multirow{3}{*}{6} &
  - &
  \multirow{3}{*}{7} &
  - \\
 &
  \(R^2\) &
  \textcolor{blue}{\bf{97.60\%}} &
  88.22\% &
   &
  83.04\% &
   &
  70.63\% &
   &
  - &
   &
  - &
   &
  - &
   &
  - \\
 &
  MAE &
  \textcolor{blue}{\bf{13.77}} &
  35.33 &
   &
  41.89 &
   &
  54.32 &
   &
  - &
   &
  - &
   &
  - &
   &
  - \\ \hline
\multirow{3}{*}{\textbf{Eastern Asia}} &
  RMSE &
  39.13 &
  56.04 &
  \multirow{3}{*}{1} &
  \textcolor{blue}{\bf{12.12}} &
  \multirow{3}{*}{1} &
  \textcolor{blue}{\bf{12.12}} &
  \multirow{3}{*}{1} &
  \textcolor{blue}{\bf{12.12}} &
  \multirow{3}{*}{2} &
  - &
  \multirow{3}{*}{2} &
  - &
  \multirow{3}{*}{2} &
  - \\
 &
  \(R^2\) &
  93.53\% &
  86.77\% &
   &
  \textcolor{blue}{\bf{98.56\%}} &
   &
  \textcolor{blue}{\bf{98.56\% }}&
   &
  \textcolor{blue}{\bf{98.56\%}} &
   &
  - &
   &
  - &
   &
  - \\
 &
  MAE &
  26.55 &
  40.20 &
   &
  \textcolor{blue}{\bf{14.56}} &
   &
  \textcolor{blue}{\bf{14.56}} &
   &
  \textcolor{blue}{\bf{14.56}} &
   &
  - &
   &
  - &
   &
  - \\ \hline
\multirow{3}{*}{\textbf{West of USA}} &
  RMSE &
  15.74 &
  52.66 &
  \multirow{3}{*}{2} &
  51.33 &
  \multirow{3}{*}{3} &
  15.10 &
  \multirow{3}{*}{3} &
  \textcolor{blue}{\bf{14.90}} &
  \multirow{3}{*}{4} &
  \textcolor{blue}{\bf{14.90}} &
  \multirow{3}{*}{5} &
  \textcolor{blue}{\bf{14.90}} &
  \multirow{3}{*}{6} &
  \textcolor{blue}{\bf{14.90}} \\
 &
  \(R^2\) &
  98.01\% &
  77.84\% &
   &
  78.52\% &
   &
  98.12\% &
   &
  \textcolor{blue}{\bf{98.22\%}} &
   &
  \textcolor{blue}{\bf{98.22\%}} &
   &
  \textcolor{blue}{\bf{98.22\%}} &
   &
  \textcolor{blue}{\bf{98.22\%}} \\
 &
  MAE &
  12.56 &
  41.15 &
   &
  40.45 &
   &
  11.55 &
   &
  \textcolor{blue}{\bf{10.52}} &
   &
  \textcolor{blue}{\bf{10.52}} &
   &
  \textcolor{blue}{\bf{10.52}} &
   &
  \textcolor{blue}{\bf{10.52}} \\ \hline
\multirow{3}{*}{\textbf{US Center}} &
  RMSE &
  15.01 &
  50.71 &
  \multirow{3}{*}{2} &
  52.53 &
  \multirow{3}{*}{3} &
  16.60 &
  \multirow{3}{*}{3} &
  \textcolor{blue}{\bf{14.13}} &
  \multirow{3}{*}{4} &
  \textcolor{blue}{\bf{14.13}} &
  \multirow{3}{*}{5} &
  \textcolor{blue}{\bf{14.13}} &
  \multirow{3}{*}{6} &
  \textcolor{blue}{\bf{14.13}} \\
 &
  \(R^2\) &
  98.15\% &
  78.96\% &
   &
  77.89\% &
   &
  97.51\% &
   &
  \textcolor{blue}{\bf{98.37\%}} &
   &
  \textcolor{blue}{\bf{98.37\%}} &
   &
  \textcolor{blue}{\bf{98.37\%}} &
   &
  \textcolor{blue}{\bf{98.37\%}} \\
 &
  MAE &
  11.31 &
  39.67 &
   &
  42.44 &
   &
  12.30 &
   &
  \textcolor{blue}{\bf{11.32}} &
   &
  \textcolor{blue}{\bf{11.32}} &
   &
  \textcolor{blue}{\bf{11.32}} &
   &
  \textcolor{blue}{\bf{11.32}} \\ \hline
\multirow{3}{*}{\textbf{West Africa}} &
  RMSE &
  18.95 &
  59.83 &
  \multirow{3}{*}{2} &
  55.78 &
  \multirow{3}{*}{3} &
  15.02 &
  \multirow{3}{*}{3} &
  \textcolor{blue}{\bf{13.63}} &
  \multirow{3}{*}{4} &
  \textcolor{blue}{\bf{13.63}} &
  \multirow{3}{*}{5} &
  \textcolor{blue}{\bf{13.63}} &
  \multirow{3}{*}{6} &
  \textcolor{blue}{\bf{13.63}} \\
 &
  \(R^2\) &
  96.87\% &
  68.82\% &
   &
  75.01\% &
   &
  98.18\% &
   &
  \textcolor{blue}{\bf{98.39\%}} &
   &
  \textcolor{blue}{\bf{98.39\%}} &
   &
  \textcolor{blue}{\bf{98.39\%}} &
   &
  \textcolor{blue}{\bf{98.39\%}} \\
 &
  MAE &
  14.85 &
  45.11 &
   &
  42.98 &
   &
  11.30 &
   &
  \textcolor{blue}{\bf{11.24}} &
   &
  \textcolor{blue}{\bf{11.24}} &
   &
  \textcolor{blue}{\bf{11.24}} &
   &
  \textcolor{blue}{\bf{11.24}} \\ \hline
\multirow{3}{*}{\textbf{North Africa}} &
  RMSE &
  18.01 &
  60.93 &
  \multirow{3}{*}{2} &
  56.92 &
  \multirow{3}{*}{3} &
  20.17 &
  \multirow{3}{*}{3} &
  \textcolor{blue}{\bf{11.19}} &
  \multirow{3}{*}{4} &
  \textcolor{blue}{\bf{11.19}} &
  \multirow{3}{*}{5} &
  \textcolor{blue}{\bf{11.19}} &
  \multirow{3}{*}{6} &
  \textcolor{blue}{\bf{11.19}} \\
 &
  \(R^2\) &
  97.28\% &
  69.99\% &
   &
  71.89\% &
   &
  96.34\% &
   &
  \textcolor{blue}{\bf{98.95\%}} &
   &
  \textcolor{blue}{\bf{98.95\%}} &
   &
  \textcolor{blue}{\bf{98.95\%}} &
   &
  \textcolor{blue}{\bf{98.95\%}} \\
 &
  MAE &
  14.72 &
  47.35 &
   &
  42.94 &
   &
  15.47 &
   &
  \textcolor{blue}{\bf{8.56}} &
   &
  \textcolor{blue}{\bf{8.56}} &
   &
 \textcolor{blue}{\bf{8.56}} &
   &
  \textcolor{blue}{\bf{8.56}} \\ \hline
\multirow{3}{*}{\textbf{Western Europe}} &
  RMSE &
  26.31 &
  74.46 &
  \multirow{3}{*}{2} &
  77.16 &
  \multirow{3}{*}{2} &
  \textcolor{blue}{\bf{14.67}} &
  \multirow{3}{*}{2} &
  \textcolor{blue}{\bf{14.67}} &
  \multirow{3}{*}{3} &
  \textcolor{blue}{\bf{14.67}} &
  \multirow{3}{*}{3} &
  21.46 &
  \multirow{3}{*}{3} &
  - \\
 &
  \(R^2\) &
  97.27\% &
  78.17\% &
   &
  73.81\% &
   &
  \textcolor{blue}{\bf{98.59\%}} &
   &
  \textcolor{blue}{\bf{98.59\%}} &
   &
  \textcolor{blue}{\bf{98.59\%}} &
   &
  98.12\% &
   &
  - \\
 &
  MAE &
  18.54 &
  48.39 &
   &
  49.27 &
   &
  \textcolor{blue}{\bf{11.43}} &
   &
  \textcolor{blue}{\bf{11.43}} &
   &
  \textcolor{blue}{\bf{11.43}} &
   &
  17.10 &
   &
  - \\ \hline
\multirow{3}{*}{\textbf{Central America}} &
  RMSE &
  15.19 &
  53.58 &
  \multirow{3}{*}{2} &
  51.44 &
  \multirow{3}{*}{3} &
  16.08 &
  \multirow{3}{*}{3} &
  \textcolor{blue}{\bf{14.23}} &
  \multirow{3}{*}{4} &
  \textcolor{blue}{\bf{14.23}} &
  \multirow{3}{*}{5} &
  \textcolor{blue}{\bf{14.23}} &
  \multirow{3}{*}{6} &
  \textcolor{blue}{\bf{14.23}} \\
 &
  \(R^2\) &
  98.14\% &
  76.85\% &
   &
  78.77\% &
   &
  97.88\% &
   &
  \textcolor{blue}{\bf{98.37\%}} &
   &
  \textcolor{blue}{\bf{98.37\%}} &
   &
  \textcolor{blue}{\bf{98.37\%}} &
   &
  \textcolor{blue}{\bf{98.37\%}} \\
 &
  MAE &
  11.88 &
  42.17 &
   &
  40.67 &
   &
  12.87 &
   &
  \textcolor{blue}{\bf{11.40}} &
   &
  \textcolor{blue}{\bf{11.40}} &
   &
  \textcolor{blue}{\bf{11.40}} &
   &
  \textcolor{blue}{\bf{11.40}} \\ \hline
\multirow{3}{*}{\textbf{South America}} &
  RMSE &
  15.88 &
  53.06 &
  \multirow{3}{*}{2} &
  49.62 &
  \multirow{3}{*}{3} &
  16.30 &
  \multirow{3}{*}{3} &
  \textcolor{blue}{\bf{12.79}} &
  \multirow{3}{*}{4} &
  \textcolor{blue}{\bf{12.79}} &
  \multirow{3}{*}{5} &
  \textcolor{blue}{\bf{12.79}} &
  \multirow{3}{*}{6} &
  \textcolor{blue}{\bf{12.79}} \\
 &
  \(R^2\) &
  97.93\% &
  76.88\% &
   &
  79.73\% &
   &
  97.33\% &
   &
  \textcolor{blue}{\bf{98.66\%}} &
   &
  \textcolor{blue}{\bf{98.66\%}} &
   &
  \textcolor{blue}{\bf{98.66\%}} &
   &
  \textcolor{blue}{\bf{98.66\%}} \\
 &
  MAE &
  12.76 &
  41.49 &
   &
  38.12 &
   &
  13.53 &
   &
  \textcolor{blue}{\bf{10.41}} &
   &
  \textcolor{blue}{\bf{10.41}} &
   &
  \textcolor{blue}{\bf{10.41}} &
   &
  \textcolor{blue}{\bf{10.41}} \\ \hline
\multirow{3}{*}{\textbf{Southern Europe}} &
  RMSE &
  34.52 &
  77.13 &
  \multirow{3}{*}{2} &
  78.54 &
  \multirow{3}{*}{2} &
  22.21 &
  \multirow{3}{*}{2} &
  22.21 &
  \multirow{3}{*}{3} &
  22.21 &
  \multirow{3}{*}{3} &
  \textcolor{blue}{\bf{19.93}} &
  \multirow{3}{*}{4} &
  - \\
 &
  \(R^2\) &
  95.61\% &
  78.06\% &
   &
  75.83\% &
   &
  98.55\% &
   &
  98.55\% &
   &
  98.55\% &
   &
  \textcolor{blue}{\bf{98.90\%}} &
   &
  - \\
 &
  MAE &
  23.63 &
  49.37 &
   &
  49.85 &
   &
  17.29 &
   &
  17.29 &
   &
  17.29 &
   &
  \textcolor{blue}{\bf{15.38}} &
   &
  - \\ \hline
\multirow{3}{*}{\textbf{East of USA}} &
  RMSE &
  14.38 &
  51.26 &
  \multirow{3}{*}{2} &
  52.63 &
  \multirow{3}{*}{3} &
  16.24 &
  \multirow{3}{*}{3} &
  \textcolor{blue}{\bf{13.73}} &
  \multirow{3}{*}{4} &
  \textcolor{blue}{\bf{13.73}} &
  \multirow{3}{*}{5} &
  \textcolor{blue}{\bf{13.73}} &
  \multirow{3}{*}{6} &
  \textcolor{blue}{\bf{13.73}} \\
 &
  \(R^2\) &
  98.31\% &
  78.71\% &
   &
  77.38\% &
   &
  97.65\% &
   &
  \textcolor{blue}{\bf{98.46\%}} &
   &
  \textcolor{blue}{\bf{98.46\%}} &
   &
  \textcolor{blue}{\bf{98.46\%}} &
   &
  \textcolor{blue}{\bf{98.46\%}} \\
 &
  MAE &
  11.77 &
  40.17 &
   &
  40.69 &
   &
  12.67 &
   &
  \textcolor{blue}{\bf{10.44}} &
   &
  \textcolor{blue}{\bf{10.44}} &
   &
  \textcolor{blue}{\bf{10.44}} &
   &
  \textcolor{blue}{\bf{10.44}} \\ \hline
\multirow{3}{*}{\textbf{South of USA}} &
  RMSE &
  15.50 &
  52.37 &
  \multirow{3}{*}{2} &
  52.78 &
  \multirow{3}{*}{3} &
  17.03 &
  \multirow{3}{*}{3} &
  \textcolor{blue}{\bf{12.65}} &
  \multirow{3}{*}{4} &
  \textcolor{blue}{\bf{12.65}} &
  \multirow{3}{*}{5} &
  \textcolor{blue}{\bf{12.65}} &
  \multirow{3}{*}{6} &
  \textcolor{blue}{\bf{12.65}} \\
 &
  \(R^2\) &
  97.97\% &
  76.68\% &
   &
  76.65 &
   &
  97.12\% &
   &
  \textcolor{blue}{\bf{98.65\%}} &
   &
  \textcolor{blue}{\bf{98.65\%}} &
   &
  \textcolor{blue}{\bf{98.65\%}} &
   &
  \textcolor{blue}{\bf{98.65\%}} \\
 &
  MAE &
  12.91 &
  40.57 &
   &
  40.36 &
   &
  13.76 &
   &
  \textcolor{blue}{\bf{10.33}} &
   &
  \textcolor{blue}{\bf{10.33}} &
   &
  \textcolor{blue}{\bf{10.33}} &
   &
  \textcolor{blue}{\bf{10.33}} \\ \bottomrule
\end{tabular}%
}
\end{table}

\renewcommand{\arraystretch}{1} % Increase row spacing

\begin{table}[ht!]
\centering
\caption{Bubble-Clustering Federated Learning with different number of clustering, ranging from 2 to 7 (Epsilon = 10)}
\label{tab:clustering2}
\resizebox{\textwidth}{!}{%
\begin{tabular}{cccccccccccccccc}
\toprule
\multirow{2}{*}{\textbf{Regions}} &
  \multirow{2}{*}{\textbf{Metrics}} &
  \multirow{2}{*}{\textbf{TF\_Local}} &
  \multirow{2}{*}{\textbf{TF\_FedAvg}} &
  \multicolumn{12}{c}{\textbf{TF\_PA-CFL}} \\ 
\cmidrule(lr){5-16} % Corrected column rule placement
  \multicolumn{2}{c}{} &  % Empty placeholders for alignment
  & & 
  \multicolumn{2}{c}{\textbf{NC =2}} &
  \multicolumn{2}{c}{\textbf{NC =3}} &
  \multicolumn{2}{c}{\textbf{NC =4}} &
  \multicolumn{2}{c}{\textbf{NC =5}} &
  \multicolumn{2}{c}{\textbf{NC =6}} &
  \multicolumn{2}{c}{\textbf{NC =7}} \\ 
\midrule

\multirow{3}{*}{\textbf{Southeast Asia}} &
  RMSE &
  31.54 &
  49.29 &
  \multirow{3}{*}{1} &
  - &
  \multirow{3}{*}{1} &
  - &
  \multirow{3}{*}{1} &
  - &
  \multirow{3}{*}{1} &
  - &
  \multirow{3}{*}{1} &
  - &
  \multirow{3}{*}{1} &
  - \\
 &
  \(R^2\) &
  \textcolor{blue}{\bf{95.05\%}} &
  28.26\% &
   &
  - &
   &
  - &
   &
  - &
   &
  - &
   &
  - &
   &
  - \\
 &
  MAE &
  \textcolor{blue}{\bf{25.60}} &
  36.38 &
   &
  - &
   &
  - &
   &
  - &
   &
  - &
   &
  - &
   &
  - \\ \hline
\multirow{3}{*}{\textbf{South Asia}} &
  RMSE &
  \textcolor{blue}{\bf{28.26}} &
  44.22 &
  \multirow{3}{*}{5} &
  - &
  \multirow{3}{*}{9} &
  - &
  \multirow{3}{*}{5} &
  - &
  \multirow{3}{*}{5} &
  - &
  \multirow{3}{*}{5} &
  - &
  \multirow{3}{*}{5} &
  - \\
 &
  \(R^2\) &
  \textcolor{blue}{\bf{95.45\%}} &
  88.93\% &
   &
  - &
   &
  - &
   &
  - &
   &
  - &
   &
  - &
   &
  - \\
 &
  MAE &
  \textcolor{blue}{\bf{23.09}} &
  34.36 &
   &
  - &
   &
  - &
   &
  - &
   &
  - &
   &
  - &
   &
  - \\ \hline
\multirow{3}{*}{\textbf{Oceania}} &
  RMSE &
  \textcolor{blue}{\bf{19.12}} &
  42.429 &
  \multirow{3}{*}{8} &
  - &
  \multirow{3}{*}{5} &
  - &
  \multirow{3}{*}{10} &
  - &
  \multirow{3}{*}{11} &
  - &
  \multirow{3}{*}{12} &
  - &
  \multirow{3}{*}{12} &
  - \\
 &
  \(R^2\) &
  \textcolor{blue}{\bf{97.60\%}} &
  88.22\% &
   &
  - &
   &
  - &
   &
  - &
   &
  - &
   &
  - &
   &
  - \\
 &
  MAE &
  \textcolor{blue}{\bf{13.77}} &
  35.33 &
   &
  - &
   &
  - &
   &
  - &
   &
  - &
   &
  - &
   &
  - \\ \hline
\multirow{3}{*}{\textbf{Eastern Asia}} &
  RMSE &
  \textcolor{blue}{\bf{39.13}} &
  56.04 &
  \multirow{3}{*}{2} &
  - &
  \multirow{3}{*}{2} &
  - &
  \multirow{3}{*}{2} &
  - &
  \multirow{3}{*}{2} &
  - &
  \multirow{3}{*}{2} &
  - &
  \multirow{3}{*}{2} &
  - \\
 &
  \(R^2\) &
  \textcolor{blue}{\bf{93.53\%}} &
  86.77\% &
   &
  - &
   &
  - &
   &
  - &
   &
  - &
   &
  - &
   &
  - \\
 &
  MAE &
  \textcolor{blue}{\bf{26.55}} &
  40.20 &
   &
  - &
   &
  - &
   &
  - &
   &
  - &
   &
  - &
   &
  - \\ \hline
\multirow{3}{*}{\textbf{West of USA}} &
  RMSE &
  15.74 &
  52.66 &
  \multirow{3}{*}{7} &
  13.91 &
  \multirow{3}{*}{8} &
  13.83 &
  \multirow{3}{*}{8} &
  12.48 &
  \multirow{3}{*}{9} &
  12.48 &
  \multirow{3}{*}{9} &
  \textcolor{blue}{\bf{12.14}} &
  \multirow{3}{*}{9} &
  14.90 \\
 &
  \(R^2\) &
  98.01\% &
  77.84\% &
   &
  98.44\% &
   &
  98.45\% &
   &
  98.75\% &
   &
  98.75\% &
   &
  \textcolor{blue}{\bf{98.81\%}} &
   &
  98.22\% \\
 &
  MAE &
  12.56 &
  41.15 &
   &
  11.09 &
   &
  10.17 &
   &
  10.01 &
   &
  10.01 &
   &
  \textcolor{blue}{\bf{9.89}} &
   &
  10.52 \\ \hline
\multirow{3}{*}{\textbf{US Center}} &
  RMSE &
  15.01 &
  50.71 &
  \multirow{3}{*}{7} &
  14.51 &
  \multirow{3}{*}{8} &
  \textcolor{blue}{\bf{11.21}} &
  \multirow{3}{*}{8} &
  11.73 &
  \multirow{3}{*}{9} &
  11.73 &
  \multirow{3}{*}{10} &
  - &
  \multirow{3}{*}{10} &
  - \\
 &
  \(R^2\) &
  98.15\% &
  78.96\% &
   &
  98.27\% &
   &
  \textcolor{blue}{\bf{98.89\%}} &
   &
  98.71\% &
   &
  98.71\% &
   &
  - &
   &
  - \\
 &
  MAE &
  11.31 &
  39.67 &
   &
  12.04 &
   &
  \textcolor{blue}{\bf{8.81}} &
   &
  9.74 &
   &
  9.74 &
   &
  - &
   &
  - \\ \hline
\multirow{3}{*}{\textbf{West Africa}} &
  RMSE &
  18.95 &
  59.83 &
  \multirow{3}{*}{7} &
  7.86 &
  \multirow{3}{*}{8} &
  9.21 &
  \multirow{3}{*}{8} &
  9.26 &
  \multirow{3}{*}{9} &
  9.26 &
  \multirow{3}{*}{9} &
  \textcolor{blue}{\bf{6.47}} &
  \multirow{3}{*}{9} &
  13.63 \\
 &
  \(R^2\) &
  96.87\% &
  68.82\% &
   &
  99.46\% &
   &
  99.31\% &
   &
  99.24\% &
   &
  99.24\% &
   &
  \textcolor{blue}{\bf{99.63\%}} &
   &
  98.39\% \\
 &
  MAE &
  14.85 &
  45.11 &
   &
  5.76 &
   &
  7.86 &
   &
  7.84 &
   &
  7.84 &
   &
  \textcolor{blue}{\bf{5.25}} &
   &
  11.24 \\ \hline
\multirow{3}{*}{\textbf{North Africa}} &
  RMSE &
  18.01 &
  60.93 &
  \multirow{3}{*}{7} &
  \textcolor{blue}{\bf{13.14}} &
  \multirow{3}{*}{7} &
  13.28 &
  \multirow{3}{*}{7} &
  13.28 &
  \multirow{3}{*}{7} &
  - &
  \multirow{3}{*}{7} &
  - &
  \multirow{3}{*}{7} &
  - \\
 &
  \(R^2\) &
  97.28\% &
  69.99\% &
   &
  \textcolor{blue}{\bf{98.45\%}} &
   &
  98.43\% &
   &
  98.43\% &
   &
  - &
   &
  - &
   &
  - \\
 &
  MAE &
  14.72 &
  47.35 &
   &
  \textcolor{blue}{\bf{10.94}} &
   &
  11.41 &
   &
  11.41 &
   &
  - &
   &
  - &
   &
  - \\ \hline
\multirow{3}{*}{\textbf{Western Europe}} &
  RMSE &
  \textcolor{blue}{\bf{26.31}} &
  74.46 &
  \multirow{3}{*}{3} &
  - &
  \multirow{3}{*}{3} &
  - &
  \multirow{3}{*}{3} &
  - &
  \multirow{3}{*}{3} &
  - &
  \multirow{3}{*}{3} &
  - &
  \multirow{3}{*}{3} &
  - \\
 &
  \(R^2\) &
  \textcolor{blue}{\bf{97.27\%}} &
  78.17\% &
   &
  - &
   &
  - &
   &
  - &
   &
  - &
   &
  - &
   &
  - \\
 &
  MAE &
  \textcolor{blue}{\bf{18.54}} &
  48.39 &
   &
  - &
   &
  - &
   &
  - &
   &
  - &
   &
  - &
   &
  - \\ \hline
\multirow{3}{*}{\textbf{Central America}} &
  RMSE &
  15.19 &
  53.58 &
  \multirow{3}{*}{6} &
  \textcolor{blue}{\bf{13,57}} &
  \multirow{3}{*}{6} &
  \textcolor{blue}{\bf{13,57}}&
  \multirow{3}{*}{6} &
  \textcolor{blue}{\bf{13,57}} &
  \multirow{3}{*}{6} &
  \textcolor{blue}{\bf{13,57}} &
  \multirow{3}{*}{6} &
  \textcolor{blue}{\bf{13,57}} &
  \multirow{3}{*}{6} &
  \textcolor{blue}{\bf{13,57}} \\
 &
  \(R^2\) &
  98.14\% &
  76.85\% &
   &
  \textcolor{blue}{\bf{98.51\%}} &
   &
  \textcolor{blue}{\bf{98.51\%}} &
   &
  \textcolor{blue}{\bf{98.51\%}}&
   &
  \textcolor{blue}{\bf{98.51\%}} &
   &
  \textcolor{blue}{\bf{98.51\%}} &
   &
  \textcolor{blue}{\bf{98.51\%}} \\
 &
  MAE &
  11.88 &
  42.17 &
   &
  \textcolor{blue}{\bf{10.87\%}} &
   &
  \textcolor{blue}{\bf{10.87\%}} &
   &
  \textcolor{blue}{\bf{10.87\%}} &
   &
  \textcolor{blue}{\bf{10.87\%}}&
   &
  \textcolor{blue}{\bf{10.87\%}}&
   &
  \textcolor{blue}{\bf{10.87\%}} \\ \hline
\multirow{3}{*}{\textbf{South America}} &
  RMSE &
  15.88 &
  53.06 &
  \multirow{3}{*}{6} &
  \textcolor{blue}{\bf{11.31\%}} &
  \multirow{3}{*}{6} &
  \textcolor{blue}{\bf{11.31\%}} &
  \multirow{3}{*}{6} &
  \textcolor{blue}{\bf{11.31\%}} &
  \multirow{3}{*}{6} &
  \textcolor{blue}{\bf{11.31\%}} &
  \multirow{3}{*}{6} &
  \textcolor{blue}{\bf{11.31\%}} &
  \multirow{3}{*}{6} &
  \textcolor{blue}{\bf{11.31\%}} \\
 &
  \(R^2\) &
  97.93\% &
  76.88\% &
   &
  \textcolor{blue}{\bf{98.95\%}} &
   &
  \textcolor{blue}{\bf{98.95\%}} &
   &
  \textcolor{blue}{\bf{98.95\%}} &
   &
  \textcolor{blue}{\bf{98.95\%}}&
   &
  \textcolor{blue}{\bf{98.95\%}} &
   &
  \textcolor{blue}{\bf{98.95\%}} \\
 &
  MAE &
  12.76 &
  41.49 &
   &
  \textcolor{blue}{\bf{9.09\%}} &
   &
  \textcolor{blue}{\bf{9.09\%}} &
   &
  \textcolor{blue}{\bf{9.09\%}} &
   &
  \textcolor{blue}{\bf{9.09\%}} &
   &
  \textcolor{blue}{\bf{9.09\%}} &
   &
  \textcolor{blue}{\bf{9.09\%}} \\ \hline
\multirow{3}{*}{\textbf{Southern Europe}} &
  RMSE &
  \textcolor{blue}{\bf{34.52}} &
  77.13 &
  \multirow{3}{*}{4} &
  - &
  \multirow{3}{*}{4} &
  - &
  \multirow{3}{*}{4} &
  - &
  \multirow{3}{*}{4} &
  - &
  \multirow{3}{*}{4} &
  - &
  \multirow{3}{*}{4} &
  - \\
 &
  \(R^2\) &
  \textcolor{blue}{\bf{95.61\%}} &
  78.06\% &
   &
  - &
   &
  - &
   &
  - &
   &
  - &
   &
  \textbf{-} &
   &
  \textbf{-} \\
 &
  MAE &
  \textcolor{blue}{\bf{23.63}} &
  49.37 &
   &
  - &
   &
  - &
   &
  - &
   &
  - &
   &
  - &
   &
  - \\ \hline
\multirow{3}{*}{\textbf{East of USA}} &
  RMSE &
  14.38 &
  51.26 &
  \multirow{3}{*}{7} &
  12.91 &
  \multirow{3}{*}{7} &
  \textcolor{blue}{\bf{10.82}} &
  \multirow{3}{*}{7} &
  \textcolor{blue}{\bf{10.82}} &
  \multirow{3}{*}{8} &
  - &
  \multirow{3}{*}{8} &
  - &
  \multirow{3}{*}{8} &
  - \\
 &
  \(R^2\) &
  98.31\% &
  78.71\% &
   &
  98.64\% &
   &
  \textcolor{blue}{\bf{99.05\%}} &
   &
  \textcolor{blue}{\bf{99.05\%}} &
   &
  - &
   &
  - &
   &
  - \\
 &
  MAE &
  11.77 &
  40.17 &
   &
  10.20 &
   &
  \textcolor{blue}{\bf{8.90}} &
   &
  \textcolor{blue}{\bf{8.90}} &
   &
  - &
   &
  - &
   &
  - \\ \hline
\multirow{3}{*}{\textbf{South of USA}} &
  RMSE &
  15.50 &
  52.37 &
  \multirow{3}{*}{7} &
  15.20 &
  \multirow{3}{*}{8} &
  \textcolor{blue}{\bf{12.08}} &
  \multirow{3}{*}{9} &
  - &
  \multirow{3}{*}{10} &
  - &
  \multirow{3}{*}{11} &
  - &
  \multirow{3}{*}{11} &
  - \\
 &
  \(R^2\) &
  97.97\% &
  76.68\% &
   &
  98.04\% &
   &
  \textcolor{blue}{\bf{98.88\%}} &
   &
  - &
   &
  - &
   &
  - &
   &
  - \\
 &
  MAE &
  12.91 &
  40.57 &
   &
  12.09 &
   &
  \textcolor{blue}{\bf{10.14}} &
   &
  - &
   &
  - &
   &
  - &
   &
  - \\ \hline
\end{tabular}%
}
\end{table}

In addition to the effects of encryption, it is crucial to examine how the number of clusters impacts the performance of the PA-CFL method. By leveraging the Davies-Bouldin Index, PA-CFL identifies the optimal number of clusters that yield the lowest score for effective client segmentation. However, as shown in \autoref{fig:score}, alternative low Davies-Bouldin scores may suggest other viable clustering configurations, highlighting diverse combinations of clients represented as bubbles. This raises the question of how clustering settings associated with each Davies-Bouldin Index influence performance. For this experiment, epsilon is set to a default value of 10, ensuring a moderate level of encryption, as clustering settings tend to stabilize when epsilon approaches 10.
The results, presented in \autoref{tab:clustering1} and \autoref{tab:clustering2}, highlight the best performance metrics (RMSE, MAE, and R²) across all methods, emphasized in bold. \autoref{tab:clustering1} illustrates how grouping settings and PA-CFL performance evolve for various retail regions as the number of clusters NC increases from 2 to 7. 
Besides, it also shows the specific group number of the clustering for each region.
When combined with \autoref{fig:score}, it becomes evident that as the number of bubbles rises from 2 to 7, the Davies-Bouldin score drops sharply from approximately 0.38 to around 0.005. Notably, when clustering clients into two bubbles, most clients do not benefit from PA-CFL, resulting in suboptimal clustering with a Davies-Bouldin score near 0.4. Similarly, with three clusters, about one-third of clients perform poorly in demand forecasting, as the Davies-Bouldin score remains high. Effective grouping is only achieved when the number of clusters reaches four, allowing all clients to benefit from PA-CFL, with the Davies-Bouldin score falling below 0.3.

Interestingly, as the number of clusters increases to 7 and the Davies-Bouldin score drops to 0.05, clients participating in PA-CFL show significant performance improvements compared to local learning. Nearly all clients achieve their best performance, reflected by the lowest RMSE and MAE, along with the highest R² values. However, as the number of clusters grows, fewer clients are selected to form bubbles in PA-CFL. \autoref{tab:clustering2} further demonstrates this trend, showing clustering numbers NC ranging from 8 to 13, with the Davies-Bouldin score rising but remaining below 0.2. These findings indicate that while the Davies-Bouldin score increases, PA-CFL performance remains robust as long as it stays under 0.2. However, increasing the number of clusters may reduce client participation, limiting the utility of PA-CFL by excluding clients who wish to join the federated learning system. Across all retail regions, most clients can join one of the PA-CFL bubbles, though Oceania is treated as an outlier and cannot participate in collaborative demand forecasting.
Thus, our PA-CFL method introduces a critical trade-off between the number of clusters and the number of clients willing to engage in federated learning. The Davies-Bouldin Index plays a pivotal role in this dynamic, significantly influencing the effectiveness of the PA-CFL system in real-world applications.

\section{Discussion and Implication}\label{sec:disscusion}

This paper provides novel insights into the application of federated learning for demand forecasting, addressing key challenges identified in prior studies. Specifically, it highlights the crucial role of clustering in federated learning to manage the heterogeneity of demand data among cross-border retailers. 
Additionally, to enhance data privacy in clients clustering, we introduce a privacy-preserving mechanism that groups potential clients into distinct bubbles before initiating federated learning.
Building on these groupings, the proposed PA-CFL framework not only enhances model performance but also provides a systematic approach to designing incentive mechanisms for FL participants. The PA-CFL algorithm enables the evaluation of data value and facilitates equitable benefit distribution among retailers engaged in collaborative demand forecasting, thereby increasing the reliability and practicality of FL applications.
Furthermore, PA-CFL can be leveraged to build a fair reward distribution system that incentivizes honest participation while imposing penalties on clients who deliberately manipulate or submit misleading data. This dual mechanism ensures both the integrity of the collaborative process and the trustworthiness of the participants, making PA-CFL a robust and scalable solution for real-world FL implementations in retail demand forecasting.
.

However, this study has certain limitations. One key limitation is the scope of experimental datasets, which should be expanded to further validate the model’s effectiveness across a broader range of heterogeneous data distributions. The current experiments, while demonstrating the efficacy of the proposed approach, may not fully capture the complexities of highly diverse retail environments. 
Additionally, in real-world scenarios, the number of retailers participating in federated learning could scale to hundreds of millions, significantly increasing computational, storage, and communication costs. Managing such large-scale participation poses challenges in model aggregation efficiency, network latency, and system scalability. 

From a practical standpoint, the proposed PA-CFL method significantly enhances demand forecasting accuracy within global supply chains while optimizing federated learning processes. The findings offer actionable insights for retailers and supply chain decision-makers seeking to leverage FL for secure data sharing and improved forecasting in complex, heterogeneous environments. Furthermore, PA-CFL’s ability to identify suitable participants enables more effective data valuation and benefit-sharing mechanisms, fostering stronger and more efficient collaboration across supply chain networks.
Beyond improving forecasting accuracy, PA-CFL also strengthens the security and robustness of federated learning systems. In real-world applications, new clients joining the system may introduce poisoned or malicious data, jeopardizing model integrity. By effectively clustering participants and filtering out unreliable inputs, PA-CFL mitigates these risks, ensuring a more resilient and trustworthy federated learning framework. Future work could further enhance these security measures by integrating adversarial detection techniques and blockchain-based verification mechanisms to reinforce data integrity.

\section{Conclusion and Future Work}\label{sec:conclusion}

This study introduces the Privacy-Adaptive Clustered Federated Learning framework, a novel approach designed to enhance retail demand forecasting by organizing heterogeneous retail data into clusters in a privacy-preserving manner. By leveraging Transformer-based models and a real-world global supply chain dataset, we demonstrate that PA-CFL effectively building sub-federated learning processes within distinct ``bubbles'' to accommodate diverse customer data distributions, resulting in significantly improved forecasting accuracy.
Moreover, our findings underscore the robustness and adaptability of PA-CFL across varying encryption levels and cluster configurations. The results indicate that PA-CFL offers a scalable and flexible FL framework that optimally balances privacy preservation, cluster efficiency, and participant diversity. Additionally, the framework mitigates risks associated with unreliable or adversarial clients, ensuring a more secure and reliable learning environment.

Future research will focus on further evaluating PA-CFL in demand forecasting by incorporating a more diverse range of datasets that exhibit higher levels of heterogeneity. Additionally, we aim to scale the framework to accommodate a significantly larger number of participants, reflecting real-world FL applications that involve millions of retailers. Further enhancements will include optimizing communication efficiency, improving computational scalability, and integrating advanced anomaly detection techniques to better identify and mitigate adversarial behavior. Moreover, extending the PA-CFL framework to other domains, such as financial forecasting and healthcare demand prediction, could provide broader insights into its applicability in privacy-sensitive and heterogeneous environments.

\newpage
%% Reference
\bibliographystyle{plainnat}
\small{
    \bibliography{reference}
}

%% Appendix
\newpage
\appendix

\end{document}